\documentclass{article}

 \usepackage[main, final]{neurips_2026}
\usepackage[utf8]{inputenc} % allow utf-8 input
\usepackage[T1]{fontenc}    % use 8-bit T1 fonts
\usepackage{hyperref}       % hyperlinks
\usepackage{url}            % simple URL typesetting
\usepackage{booktabs}       % professional-quality tables
\usepackage{amsfonts}       % blackboard math symbols
\usepackage{nicefrac}       % compact symbols for 1/2, etc.
\usepackage{microtype}      % microtypography
\usepackage{xcolor}         % colors
\usepackage{wrapfig}
\usepackage{kotex}
\usepackage{setspace}
\usepackage{wrapfig}
\usepackage{amsmath}
\usepackage{amssymb}
\usepackage{mathtools}
\usepackage{amsthm}
\usepackage[capitalise]{cleveref}
\usepackage[ruled,vlined]{algorithm2e}
\usepackage{thmtools,thm-restate}
\usepackage{cancel}
\usepackage{pifont}
\usepackage{algorithmic}
\usepackage{multicol,multirow}
\usepackage{subcaption}
\usepackage{makecell}
\usepackage{float}
\usepackage{booktabs}
\usepackage{graphicx}
\usepackage{svg}
\title{ASAP: Attention Sink Anchored Pruning}

\author{%
  Jaehyuk Lee \\
  Department of Mathematics \\
  Korea University \\
  Seoul, Republic of Korea \\
  \texttt{jaehyeokbear@korea.ac.kr} \\
  \And
  Hanyoung Kim \\
  Department of Mathematics \\
  Korea University \\
  Seoul, Republic of Korea \\
  \texttt{pinga999@korea.ac.kr} \\
  \AND
  Yanggee Kim \\
  Korea University \\
  Seoul, Republic of Korea \\
  \texttt{wheresmadog@korea.ac.kr} \\
  \And
  Donghun Lee \thanks{Corresponding author.} \\
  Department of Mathematics \\
  Korea University \\
  Seoul, Republic of Korea \\
  \texttt{holy@korea.ac.kr} \\
}

\begin{document}

\maketitle

\begin{abstract}
Vision Transformers (ViTs) face severe computational bottlenecks due to the quadratic complexity of self-attention at high resolutions. Existing token reduction methods rely on local metrics—such as single-layer attention scores—that are inherently vulnerable to the \emph{attention sink} phenomenon, where uninformative tokens are paradoxically preserved over salient foreground objects. We propose \textbf{ASAP} (Attention Sink Anchored Pruning), a training-free framework that recasts this sink as a feature. Modeling ViT information flow as a Lazy Random Walk, ASAP identifies the sink as a dominant accumulator of probability mass. By computing the \emph{diffusion distance} to the sink within the cumulative transition matrix, ASAP partitions tokens via \emph{Radial Diffusion Clustering} and compresses background redundancy through \emph{Transition Weight Pooling} in a single shot. Extensive experiments across image, video, and vision-language tasks demonstrate ASAP outperforms state-of-the-art methods, accelerating throughput by up to 48\% while maintaining or even exceeding baseline accuracy. Code will be available at \url{https://github.com/2jaehyuk/Attention-Sink-Anchored-Pruning}.

\end{abstract}

\section{Introduction}
Vision Transformers (ViTs) have demonstrated remarkable performance across a wide range of visual understanding tasks, forming the backbone of modern Visual Language Models (VLMs) and video transformers. However, as input resolution and sequence length scale up, the number of visual tokens $N$ grows rapidly, leading to severe computational and memory bottlenecks due to the quadratic complexity of self-attention. To mitigate this, numerous token pruning and merging methods have been proposed.

Existing approaches such as ToMe \cite{bolya2023token}, FastV \cite{chen2024image}, PPT \cite{wu2023ppt}, and VisPruner \cite{zhang2025beyond} predominantly rely on local metrics, single-layer attention scores or pairwise cosine similarities, to identify redundant tokens. These metrics fail to capture the nonlinear, global structure of the data manifold that emerges as the network deepens. As a result, semantically important but low-attention tokens (e.g., small foreground objects) are frequently discarded alongside background tokens, causing irreversible information loss. 
% A compounding issue is the well-known attention sink phenomenon \cite{darcet2024vision},
This is exacerbated by the attention sink phenomenon, wherein disproportionate attention mass concentrates on semantically irrelevant patches. Since existing methods equate large attention scores with importance, they remain fundamentally vulnerable to this artifact.

Recent work such as Zero-TPrune \cite{wang2024zero} attempts to go beyond local metrics by leveraging the stationary distribution of per-layer transition matrices. However, computing stationary distributions independently at each layer does not escape the local information bottleneck, and the iterative convergence procedure incurs non-trivial computational overhead.

In this work, we inherit the Markov chain interpretation of Attention Rollout \cite{abnar2020quantifying}, but depart from prior work in a critical way: whereas existing methods extract scalar importance scores from accumulated attention, we treat the cumulative transition matrix $P^{(t)}$ as defining a \textbf{geometric space} and measure token relationships via \textbf{diffusion distance} within this space. We reframe the attention sink not merely as a model artifact but as a token that behaves as a mass-accumulating attractor in our Markov chain formulation, a gravity well toward which redundant background information naturally converges, and exploit it as the geometric anchor for token clustering.

\begin{figure}[t]
    \centering
    \includegraphics[width=\textwidth]{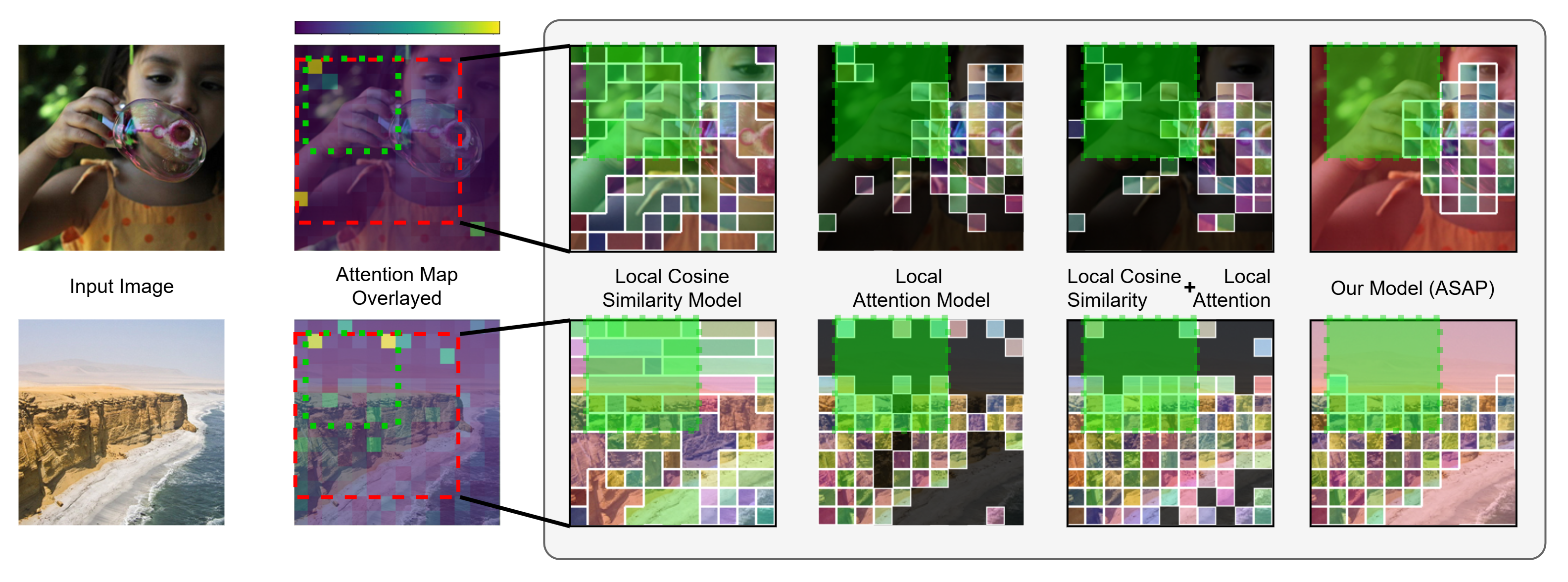}
    \caption{Visual comparison of token reduction under a fixed budget. Local cosine similarity (ToMe\cite{bolya2023token}), local attention score (FastV\cite{chen2024image}, EViT\cite{liang2022not}), and their combination (PPT\cite{wu2023ppt}) all fail to reliably separate foreground from background due to attention sink interference. Our method (ASAP) explicitly exploits the sink (red dashed) as a geometric anchor, consistently preserving key foreground objects (green dotted).}\label{fig:existing_vs_ours}
\end{figure}

Building on this view, we propose ASAP (Attention Sink Anchored Pruning), an efficient token reduction framework that performs a single clustering operation at the layer where the attention sink emerges (Figure~\ref{fig:existing_vs_ours}). Rather than relying on fragile local metrics, ASAP leverages the diffusion distance to the sink as a single criterion for token reduction, compressing background redundancy while preserving semantically important foreground structures.

We summarize our main contributions as follows:
\begin{itemize}
    \item \textbf{Modeling the attention sink as a geometric anchor.} We model ViT information flow as a Lazy Random Walk and treat the attention sink as a mass-accumulating attractor of a Markov chain, leveraging a known artifact as an organizing structure for token reduction.
    
    \item \textbf{Efficient single-shot token reduction via diffusion distance.} We compute $N$ diffusion distances to the sink in $\mathcal{O}(N^2)$ on the cumulative transition matrix $P^{(t^*)}$, followed by $\mathcal{O}(N\log N)$ cluster assignment and Transition Weight Pooling---requiring no K-Means, FPS, or manual layer selection (the reduction layer $t^*$ is determined dynamically).
    
    \item \textbf{State-of-the-art training-free performance across three domains.} Without fine-tuning the backbone, ASAP achieves the best average accuracy retention among training-free methods on ImageNet-1K, Kinetics-400, and LLaVA.
    
\end{itemize}

\section{Related Works}
\subsection{Token Pruning and Merging}

Token reduction methods for ViTs fall into two paradigms.
\emph{Pruning} methods discard tokens deemed unimportant: DynamicViT~\cite{rao2021dynamicvit} and EViT~\cite{liang2022not} train lightweight gating modules, while training-free alternatives such as FastV~\cite{chen2024image}, SparseVLM~\cite{zhang2024sparsevlm}, and VisPruner~\cite{zhang2025beyond} rank tokens by attention scores at a predetermined layer.
\emph{Merging} methods combine redundant tokens: ToMe~\cite{bolya2023token} progressively matches similar pairs via bipartite matching on key vectors, and PPT~\cite{wu2023ppt} adaptively switches between pruning and pooling.
PruneSID~\cite{fang2026prune} clusters tokens via PCA-based grouping and removes intra-group redundancy through NMS, but still operates on static single-layer embeddings.
All of these approaches rely on local, single-layer signals, making them susceptible to the attention sink: tokens absorbing disproportionate attention are incorrectly retained or used as merging anchors.

In contrast, ASAP makes a single reduction decision from the cumulative transition matrix $P^{(t^*)}$, partitioning tokens via scalar distance to the sink in $\mathcal{O}(N^2)$ rather than through repeated $\mathcal{O}(N^2d)$ pairwise comparisons, SVD, or explicit importance--diversity balancing; its dominant cost, the accumulation of $P^{(t^*)}$, is incurred once per forward pass (Table~\ref{tab:complexity}).

\subsection{Attention Graph-based Methods}\label{sec:relwork2}

Attention Rollout \cite{abnar2020quantifying} accumulates per-layer attention matrices as $\tilde{A}^1 \times \cdots \times \tilde{A}^t$ ($\tilde{A}^l = 0.5 A^l + 0.5 I$ to account for residual connections) to track global information flow, originally as a post-hoc interpretability tool. Zero-TPrune \cite{wang2024zero} applies Weighted PageRank to per-layer attention graphs but computes stationary distributions independently at each layer, without escaping the local information bottleneck. Rollout-Guided Token Pruning \cite{dinai2025rollout} propagates rolled-out scores across video frames but reduces the cumulative transition matrix to a scalar ranking signal.
% In contrast, our method treats $A^t$ as defining a \textbf{metric space} and computes diffusion distances within this space, rather than extracting scalar importance scores. Furthermore, we identify the attention sink as the absorbing state of the Markov chain and use it as a geometric anchor for a single-shot clustering operation — eliminating iterative per-layer computation entirely.
In contrast, our method treats $P^{(t)}$ as defining a metric space and computes diffusion distances within it, using the attention sink as a geometric anchor for a single-shot clustering operation.

\subsection{Attention Sink in Vision Transformers}\label{subsec:attn_sink}

The attention sink phenomenon—where disproportionate attention concentrates on uninformative tokens—emerges in language~\cite{xiao2023efficient, gu2024attention} and vision~\cite{darcet2024vision} transformers. Recent work reframes them as functional structures preventing representational collapse~\cite{barbero2025llms} or acting as coordinate anchors~\cite{ruscio2025you}.
Crucially, \cite{darcet2024vision} quantitatively demonstrates these outlier tokens consistently form at redundant background regions of high neighbor similarity, aggregating global features with minimal local spatial information. This confirms the attention sink is semantically uninformative yet structurally predictable in its spatial distribution.

Building on this, we repurpose the attention sink from a computational obstacle into our primary organizing signal. We achieve this by modeling the sink as a Markov chain mass-accumulating attractor and using token-to-sink diffusion distance to measure semantic relevance.

\section{Methods}

\begin{figure}[t]
    \centering
    \includegraphics[width=\linewidth]{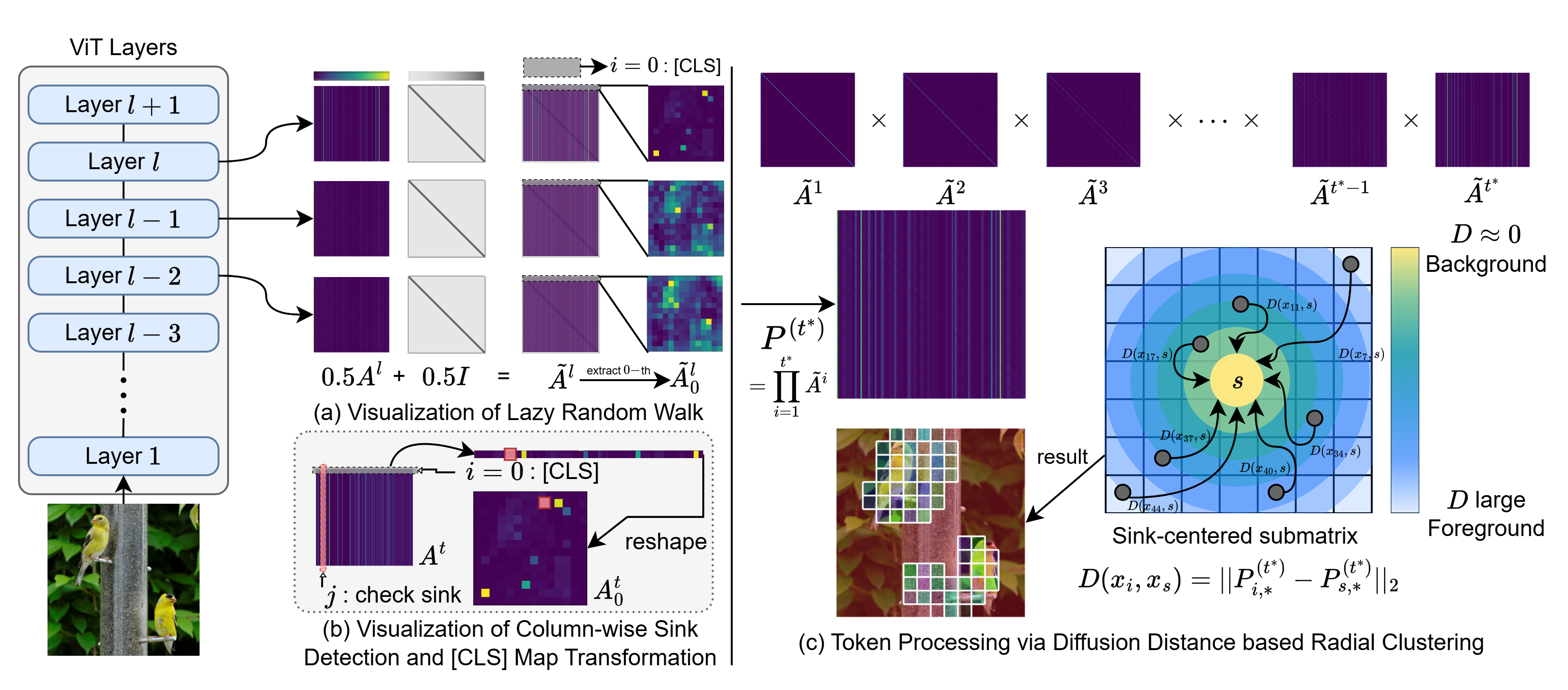} 
    \caption{Overview of ASAP. (a) Lazy Random Walk models ViT information flow for stable convergence. (b) Dynamic sink detection identifies emergence time $t^*$ via a column-sum trigger on $P^{(t)}$. (c) Radial Diffusion Clustering partitions tokens into semantic sets using diffusion distance $D(x_i, x_s)$ to the sink in $P^{(t^*)}$ space.}
    \label{fig:main_figure}
    \vspace{-4mm}
\end{figure}

ASAP leverages global information flow and the attention sink's geometric properties. The framework operates in three stages (Figure \ref{fig:main_figure}): (a) formulating ViT information flow as a Lazy Random Walk (\cref{sec:lazy}), (b) dynamically detecting the attention sink to define the diffusion distance geometry within the cumulative transition matrix (\cref{sec:sink}), and (c) partitioning tokens via Radial Diffusion Clustering and executing token reduction through pooling and optional plug-in pruning (\cref{sec:diffusiondist}).

\subsection{Information Flow as Lazy Random Walk}\label{sec:lazy}

The self-attention map $A^l \in \mathbb{R}^{N \times N}$ of a Vision Transformer satisfies $\sum_j A^l_{i,j} = 1$ for all $i$ due to the Softmax normalization, and can therefore be interpreted as a probability transition matrix of a Markov chain over the patch graph \cite{abnar2020quantifying}. 
To mathematically incorporate the residual connection—which the attention matrix $A^l$ alone cannot capture—as a tendency to preserve token identity across layers (Figure~\ref{fig:main_figure} (a)), we define the effective transition matrix as a Lazy Random Walk:

\begin{equation}
    \tilde{A}^l = \alpha A^l + (1 - \alpha) I
    \label{eq:lazy_rw}
\end{equation}
where $A^l = \frac{1}{H}\sum_{h=1}^{H} A^{l,h}$ is the head-averaged attention map at layer $l$ following \citep{abnar2020quantifying}, $I$ is the identity matrix, and $\alpha \in (0,1)$ balances the information gathered via attention with the information preserved via the residual connection.

The identity term ensures all diagonal entries of $\tilde{A}^l$ are strictly positive, guaranteeing aperiodicity by the Perron-Frobenius theorem~\cite{horn2012matrix} and stable convergence without oscillation. The cumulative transition matrix after $t$ layers is defined as the sequential matrix product:
\begin{equation}
P^{(t)} = \tilde{A}^1 \times \tilde{A}^2 \times \cdots \times \tilde{A}^t
\end{equation}
The $i$-th row $P^{(t)}_{i,*}$ represents the probability distribution over all tokens that information originating from token $x_i$ reaches after passing through $t$ layers. Unlike the local attention pattern of a single layer, $P^{(t)}$ encodes the global information flow across the entire network, and serves as the manifold space in which diffusion distances are subsequently computed.

\subsection{Sink Emergence and Diffusion Distance Geometry}\label{sec:sink}

As discussed in \cref{subsec:attn_sink}, attention sinks predominantly emerge at semantically neutral background regions across diverse architectures. In our Markov chain formulation, the sink corresponds to a mass-accumulating attractor: once it emerges, it accumulates monotonically increasing probability mass (Appendix~\ref{appendix:proof1}). Rather than treating this convergence as an artifact to suppress, we exploit it to define the geometry of $P^{(t^*)}$. We detect sink emergence via a column-sum threshold $\tau$, which we find robust across architectures (\cref{subsec:hyperparmas}; Appendix~\ref{appendix:sink_emergence}), and identify the sink location $s$ as:

\begin{equation}
    t^* = \min \left\{ t \;\middle|\; 
    \max_{j \geq 1} \sum_{i=1}^{N} P^{(t)}_{i,j} > \tau \right\},
    \quad 
    x_s = \arg\max_{j \geq 1} \, P^{(t^*)}_{0,\, j}
    \label{eq:sink_detection}
\end{equation}

Because the sink acts as a mass-accumulating attractor for all tokens, the argmax of the [CLS] token's attention ($P_{0,*}^{(t^*)}$) serves as a reliable and computationally efficient proxy for the sink index. Terminating the accumulation at $t^*$ allows the sink to fully emerge while preventing complete manifold collapse, which maintains the geometric diversity necessary for computing diffusion distances. Empirically, these distances decrease as the network deepens, a behavior consistent with the theoretical convergence of the Lazy Random Walk.

\paragraph{Diffusion Distance.}
Existing token reduction methods require either $O(N^2d)$ pairwise distance computations or expensive iterative clustering to measure inter-token similarity. We depart from this by defining the distance between an arbitrary token $x_i$ and the attention sink $x_s$ as the \textbf{diffusion distance} in the $P^{(t^*)}$ space:

\begin{equation}
    D(x_i,\, x_s) = \left\| P^{(t^*)}_{i,\,*} 
    - P^{(t^*)}_{s,\,*} \right\|_2
    \label{eq:diffusion_dist}
\end{equation}

As depicted in Figure \ref{fig:main_figure} (c), this formulation allows us to use the sink as a geometric anchor to partition tokens into semantic level sets.

Under a uniform stationary measure, diffusion distance reduces to the unweighted $\ell_2$ norm~\cite{coifman2006diffusion}, enabling efficient clustering without eigendecomposition (see Appendix~\ref{appendix:diffusion_prelim} for a formal review). Specifically, computing $N$ diffusion distances against the single sink reference requires $O(N^2)$ operations—a factor of $d$ cheaper than $O(N^2 d)$ pairwise methods and orders of magnitude below the $O(N^3)$ cost of exact eigendecomposition—followed by $O(N \log N)$ sorting for cluster assignment. Unlike local metrics, diffusion distance integrates multi-path connections on the manifold: $D(x_i, x_s) \approx 0$ implies $x_i$ routes information identically to the sink (background), while large $D(x_i, x_s)$ indicates distinct information trajectories (see Appendix~\ref{appendix:propo2}).

\subsection{Sink-Anchored Token Partitioning and Compression}\label{sec:diffusiondist}

\begin{figure}[t]
    \centering
    % === 첫 번째 줄 (Row 1) ===
    \begin{subfigure}[b]{0.48\linewidth}
        \centering
        \includegraphics[width=\linewidth]{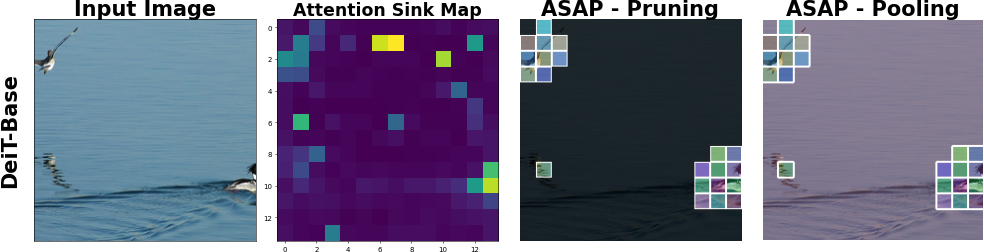}
    \end{subfigure}
    \hfill
    \begin{subfigure}[b]{0.48\linewidth}
        \centering
        \includegraphics[width=\linewidth]{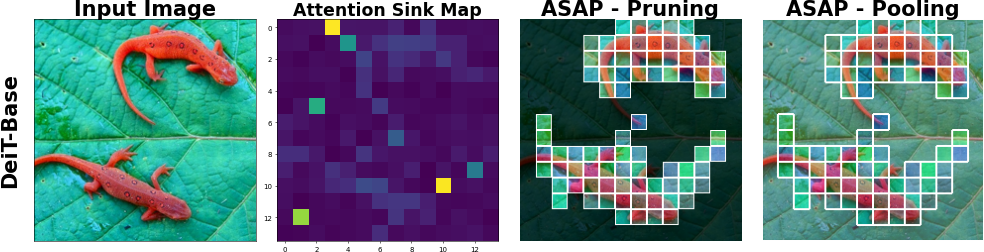}
    \end{subfigure} 
    \begin{subfigure}[b]{0.48\linewidth}
        \centering
        \includegraphics[width=\linewidth]{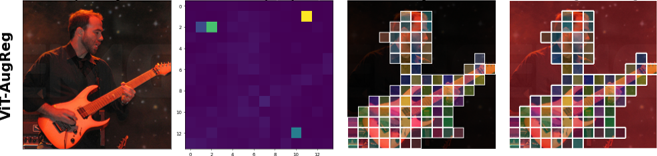} 
    \end{subfigure}
    \hfill
    \begin{subfigure}[b]{0.48\linewidth}
        \centering
        \includegraphics[width=\linewidth]{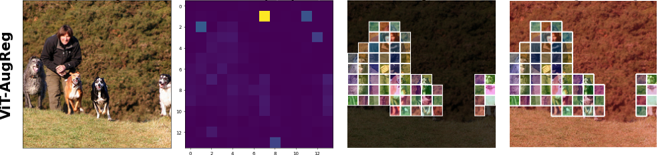}
    \end{subfigure}
    \begin{subfigure}[b]{0.48\linewidth}
        \centering
        \includegraphics[width=\linewidth]{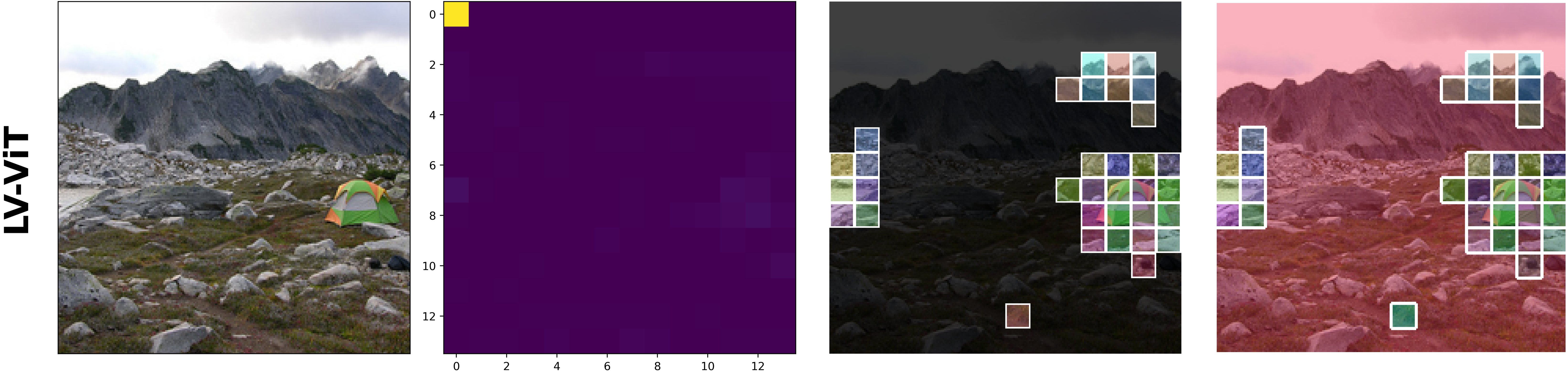} 
    \end{subfigure}
    \hfill
    \begin{subfigure}[b]{0.48\linewidth}
        \centering
        \includegraphics[width=\linewidth]{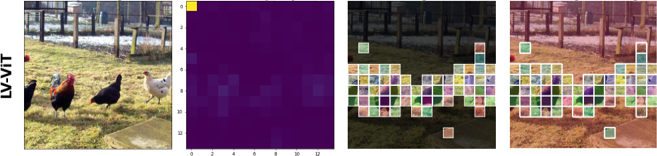}
    \end{subfigure}
    \caption{
        \textbf{Qualitative results across different backbones} (DeiT-Base, ViT-AugReg, LV-ViT-S). Our method consistently preserves foreground objects across diverse architectures and token densities.
    }
    \label{fig:sink_analysis_grid}
\end{figure}

\paragraph{Radial Diffusion Clustering.}

Using the distances derived from Eq. \ref{eq:diffusion_dist}, we propose Radial Diffusion Clustering. We first normalize the diffusion distances to $[0, 1]$ and scale them by the target cluster count $K$. Tokens are then partitioned into $K$ semantic level sets based on these scaled distances:

\begin{equation}
    C_k = \left\{\, x_i \;\middle|\; 
    k-1 \leq K \cdot \tilde{D}(x_i,\, x_s) < k \,\right\}, 
    \quad k = 1, \dots, K
    \label{eq:clustering}
\end{equation}

where $\tilde{D}(x_i, x_s)$ is the normalized diffusion distance, with the maximum-distance token assigned to $C_K$. This scaling leverages the non-uniform token distribution across the diffusion manifold, naturally achieving adaptive token pruning without complex density estimation (see Section \ref{subsec:hyperparmas} for a detailed analysis of the cluster count $K$). Consequently, $C_1$ densely groups redundant background tokens for compression, while upper clusters dynamically preserve distinct foreground structures.

\paragraph{Transition Weight Pooling.}
We compress the sink-proximal cluster $C_1$ into a single representative token via weighted pooling, where each token's feature $\mathbf{f}_i$ is weighted by its diffusion distance to the sink via softmax normalization (We set $p=1$ in experiments; Algorithm \ref{alg:asap_core} supports general $p \geq 1$):
\begin{equation}
    \mathbf{m}_1 = \sum_{x_i \in C_1} w_i\, \mathbf{f}_i, \quad 
    \text{where} \quad w_i = \frac{\exp(D(x_i,\, x_s))}
    {\sum_{x_j \in C_1} \exp(D(x_j,\, x_s))}
    \label{eq:twp}
\end{equation}

where $\mathbf{f}_i$ is the feature vector of token $x_i$. 
Tokens farther from the sink retain stronger semantic identity and thus receive higher weights, preserving residual semantic context. Surviving foreground clusters ($C_2, \dots, C_K$) are retained as-is, or optionally compressed under extreme budgets (e.g., 64 tokens in VLMs) by seamlessly integrating existing reduction methods. For instance, applying bipartite matching within the $P^{(t^*)}$ geometric space efficiently eliminates redundancy among the foreground.

\paragraph{Overall Pipeline and Implementation.} 
To summarize, the core ASAP framework executes in a single forward pass at the early layer $t^*$, consisting of: (1) identifying the sink anchor via the cumulative transition matrix, (2) partitioning tokens via Radial Diffusion Clustering, and (3) compressing the background via Transition Weight Pooling. When extreme compression is needed, the optional plug-in pruning stage is applied immediately after step (3). Detailed pseudocode for both the core pipeline (Algorithm \ref{alg:asap_core}) and the hybrid plug-in stage (Algorithm \ref{alg:asap_hybrid}) is provided in Appendix \ref{appendix:details}.

\begin{wraptable}{r}{0.46\linewidth}
\vspace{-12pt}
    \centering
    \small
    \setlength{\tabcolsep}{4pt}
    \caption{Cost added by ASAP. Accumulation runs once per forward pass.}
    \label{tab:complexity}
    \begin{tabular}{lc}
    \toprule
    Stage & Cost \\
    \midrule
    Accumulation of $P^{(t^*)}$ & $\mathcal{O}(t^* N^3)$ \\
    Diffusion distances & $\mathcal{O}(N^2)$ \\
    Clustering (sort + bin) & $\mathcal{O}(N \log N)$ \\
    \midrule
    \textit{Ref.:} one attention layer & $\mathcal{O}(N^2 d)$ \\
    \bottomrule
    \end{tabular}
\vspace{-10pt}
\end{wraptable}

\paragraph{Computational Cost.}
Table~\ref{tab:complexity} summarizes the cost ASAP adds to the backbone forward pass.
The accumulation of $P^{(t^*)}$ ($t^*{-}1$ matrix multiplications) dominates asymptotically, whereas the diffusion distances and the clustering are lower-order.
In all our settings $N \le 577$ per image or crop and $N < 2d$, so a single accumulation step, costing $2N^3$ FLOPs, is cheaper than the $4N^2d$ FLOPs of the $QK^\top$ and $AV$ products in one attention layer, and accumulation stops at $t^* < L$.
Measured on an A100, one accumulation step stays below one attention layer up to $N{=}4096$ even in strict fp32 (22\% of a layer under TF32), and the full accumulation adds only 8\,ms on LLaVA-NeXT ($N{=}2880$) (Appendix~\ref{app:scaling}).

\section{Experiments}

We evaluate ASAP across image classification, video understanding, and Vision-Language Models (VLMs). The token reduction is training-free in all experiments; the only trained component is a linear evaluation head on the frozen backbone for Kinetics-400 (Section~\ref{subsec:video}). For fair comparison, we benchmark exclusively against existing training-free token reduction methods. FLOPs and throughput are measured on a single RTX 4090 GPU, except for VLM experiments which utilize an A100 GPU. We analyze the framework's robustness through ablation studies on key hyperparameters ($K, \tau, \alpha$) in Section \ref{subsec:hyperparmas}, and detail the negligible computational overhead of the Lazy Random Walk accumulation in Appendix~\ref{appendix:overhead}.

\begin{figure}[t]
    \centering
    % === 첫 번째 줄 (Row 1) ===
    \begin{subfigure}[b]{0.48\linewidth}
        \centering
        \includegraphics[width=\linewidth]{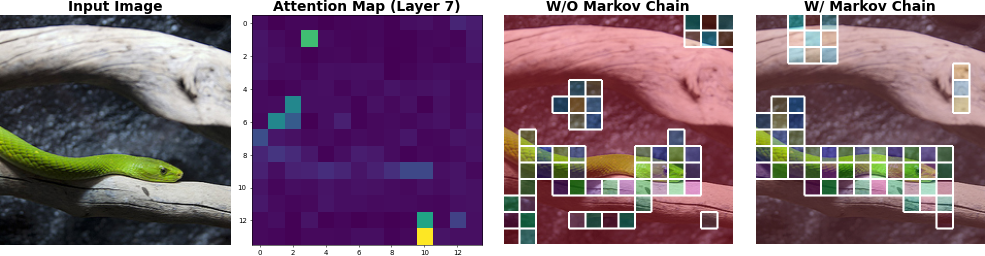}
    \end{subfigure}
    \hfill
    \begin{subfigure}[b]{0.48\linewidth}
        \centering
        \includegraphics[width=\linewidth]{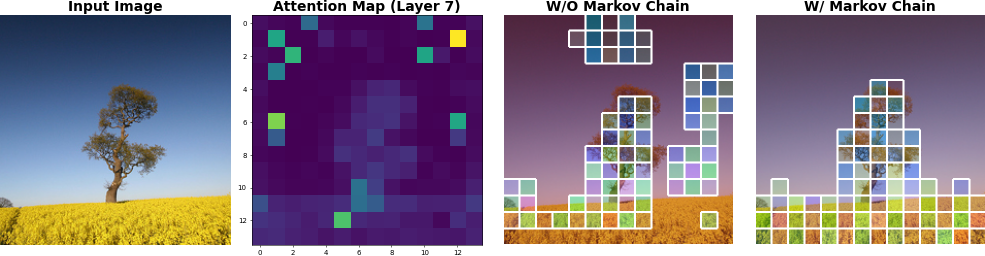}
    \end{subfigure}
    \caption{
        Necessity of cumulative attention. While our full framework (W/ Markov Chain) successfully preserves key foreground objects—such as the rapeseed flowers and the snake—the local variant (W/O Markov Chain) fails to separate them from the background.}
    \label{fig:accumul_compare}
    \vspace{-4mm}
\end{figure}

\subsection{Image Classification}\label{subsec:Image}

For image classification, we evaluate our method on ImageNet-1K~\cite{ILSVRC15_imagenet} using various backbones, including DeiT \cite{touvron2021training}, ViT-AugReg \cite{steiner2021train}, and LV-ViT-S \cite{NEURIPS2021_9a49a25d}, and compare it against existing state-of-the-art methods.
Table~\ref{tab:vit_results} shows that our method achieves the smallest accuracy drop on every backbone under comparable FLOPs budgets.
On DeiT-Base, the pooling variant achieves a 0.12\% accuracy drop with +48\% throughput, and similar trends hold on LV-ViT and ViT-AugReg.

Figure~\ref{fig:sink_analysis_grid} shows our method reliably preserves foregrounds across diverse scenes---spatially separated objects, overlapping instances, fine-grained textures---using only diffusion distance to the sink. See Appendix~\ref{appendix:additional_plots} for additional results.

To verify the necessity of cumulative attention, we evaluate a variant using only the single-layer attention matrix at $t^{*}$. As shown in Figure \ref{fig:accumul_compare}, lacking global context, this local variant fails to preserve key foreground objects and suffers a $-2.37\%$ accuracy drop on DeiT-Base. This validates that the cumulative matrix $P^{(t^*)}$ is essential for reliable foreground-background separation (Table \ref{tab:vit_results}). 

Table \ref{tab:anchor} highlights the critical role of using the attention sink as the geometric anchor. Randomly selecting an anchor can inadvertently target semantically important objects, leading to a performance drop of nearly $4-9\%$ depending on architecture compared to our proposed method. Visualizations of these failure cases with random anchors are provided in the Appendix ~\ref{appendix:random_anchor}. This advantage holds in VLMs as well: the sink outperforms both random and PCA-center anchors on LLaVA-NeXT (Appendix~\ref{appendix:anchor_selection_llavanext}).

\begin{table}[t]
    \centering
    \begin{minipage}[c]{0.55\linewidth}
        \centering
        \caption{DeiT Results on ImageNet-1k}
        \label{tab:vit_results}
        \resizebox{\linewidth}{!}{
            \begin{tabular}{llccccc}
            \toprule
            \multirow{2.5}{*}{Model} & \multirow{2.5}{*}{Method}  & \multicolumn{3}{c}{Acc($\uparrow$\%)} & \multirow{2.5}{*}{FLOPs$\downarrow$(\%)} & \multirow{2.5}{*}{Throughput$\uparrow$(\%)} \\
            \cmidrule(lr){3-5}
            & & Base. & Prun. & $\Delta$ & & \\
            \midrule
            \multirow{6}{*}{DeiT-Base}
            & DynamicViT \citep{rao2021dynamicvit}  & 81.80 & 81.30 & -0.50 & -34.09 & +38.01\ \\
            & ToMe  \citep{bolya2023token}       & 81.80 & 80.68 & -1.12 & -34.09 & +38.01 \\
            & PPT   \citep{wu2023ppt}       & 81.80 & 80.30 & -1.50 & -32.95 & +12.61 \\
            & \textbf{Ours (prun.)} & 81.80 & 81.33 & -0.47 & -36.36 & \textbf{+48.45} \\
            & \textbf{Ours (w/o accumulation)} & 81.80 & 79.43 & -2.37 & -34.13 & +43.74 \\
            & \textbf{Ours (pool.)} & 81.80 & \textbf{81.68} & \textbf{-0.12} & -35.23 & +48.21 \\
            \midrule
            \multirow{5}{*}{DeiT-Small}
            %& DynamicViT   & 79.80 & 79.30 & -0.50 & -39.13 & +27.78 \\
            & ToMe   \citep{bolya2023token}      & 79.80 & 79.10 & -0.70 & \textbf{-39.13} & \textbf{+27.78} \\
            & PPT   \citep{wu2023ppt}      & 79.80 & 79.24 & -0.56 & \textbf{-39.13} & +20.22 \\
            & ATS   \cite{fayyaz2022adaptive}        & 79.80 & 79.20 & -0.60 & -34.78 & +25.67 \\
            & ZeroTPrune \citep{wang2024zero}  & 79.80 & 79.40 & -0.40 & -34.78 & +26.89 \\
            & \textbf{Ours}         & 79.80 & \textbf{79.67} & \textbf{-0.13} & \textbf{-39.13} & +20.88 \\
            \midrule
            \multirow{4}{*}{DeiT-Tiny}
            %& DynamicViT   & 72.20 & 71.40 & -0.80 & -38.46 & +17.58 \\
            & ToMe  \citep{bolya2023token}       & 72.20 & 71.15 & -1.05 & -38.46 & +17.58 \\
            & PPT    \citep{wu2023ppt}      & 72.20 & 71.58 & -0.62 & -38.46 & +10.30 \\
            & ZeroTPrune \citep{wang2024zero}  & 72.20 & 70.40 & -1.80 & -30.77 & - \\
            & Ours         & 72.20 & 71.63 & -0.57 & -38.46 & +14.98 \\
            \midrule
            \multirow{4}{*}{ViT-AugReg}
            & ToMe   \citep{bolya2023token}      & 81.41 & 79.30 & -2.11 & -38.54 & \textbf{+21.43} \\
            & ATS   \cite{fayyaz2022adaptive}        & 81.41 & 79.21 & -2.20 & -38.90 & - \\
            & ZeroTPrune  \citep{wang2024zero} & 81.41 & 80.22 & -1.19 & -38.68 & - \\
            & Ours         & 81.41 & \textbf{80.98} & \textbf{-0.43} & \textbf{-39.12} & +19.37 \\
            \midrule
            \multirow{4}{*}{LV-ViT-S}
            & ToMe   \citep{bolya2023token}      & 83.35 & 79.80 & -3.55 & \textbf{-46.97} & +34.28 \\
            & ATS \cite{fayyaz2022adaptive}          & 83.35 & 80.40 & -2.95 & -45.45 & - \\
            & ZeroTPrune \citep{wang2024zero}  & 83.35 & 81.50 & -1.85 & \textbf{-46.97} & - \\
            & Ours         & 83.35 & \textbf{82.57} & \textbf{-0.78} & \textbf{-46.97} & \textbf{+35.66} \\
        \bottomrule
    \end{tabular}
    }
    \end{minipage}
    \hfill
    \begin{minipage}[c]{0.42\linewidth}
        \centering
        \caption{Effect of anchor selection.}
        \label{tab:anchor}
        \resizebox{\linewidth}{!}{
            \begin{tabular}{lccc}
                \toprule
                Model & Anchored & Acc$\uparrow$(\%) & Throughput$\uparrow$(\%) \\
                \midrule
                \multirow{2}{*}{DeiT-Base}
                & Random & 77.64 & +39.11 \\
                & \textbf{Attention Sink} & \textbf{81.68} & \textbf{+48.21} \\
                \midrule
                \multirow{2}{*}{LV-ViT-S}
                & Random & 73.61 & +35.22 \\
                & \textbf{Attention Sink} & \textbf{82.57} & \textbf{+35.66} \\
                \bottomrule
            \end{tabular}
        }

        \caption{Video Results on Kinetics-400}
        \label{tab:video-results}
        \resizebox{\linewidth}{!}{
            \begin{tabular}{llccccc}
            \toprule
            \multirow{2.5}{*}{Model} & \multirow{2.5}{*}{Method}  & \multicolumn{3}{c}{Acc($\uparrow$\%)} & \multirow{2.5}{*}{FLOPs$\downarrow$(\%)} & \multirow{2.5}{*}{Throughput$\uparrow$(\%)} \\
            \cmidrule(lr){3-5}
            & & Base. & Prun. & $\Delta$ & & \\
            \midrule
            \multirow{4}{*}{ClipViT-Large}
            & ToMe         & 80.78 & 78.18 & -2.60 & -34.09 & +20.0 \\
            & PPT          & 80.78 & 75.93 & -4.85 & -32.95 & +18.9 \\
            & Ours (prun.) & 80.78 & 80.13 & -0.65 & \textbf{-36.36} & \textbf{+24.5} \\
            & \textbf{Ours (pool.)} & 80.78 & \textbf{80.57} & \textbf{-0.21} & -35.23 & +23.7 \\
        \bottomrule
    \end{tabular}
    }
        \caption{End-to-End Latency Comparison.}
        \label{tab:latency_comparison}
        \resizebox{\linewidth}{!}{
        \begin{tabular}{lcccc}
            \toprule
            Method & Tokens & Avg Inference (ms) & Prefill (ms) & Acc (\%) \\
            \midrule
            LLaVA (Full)         & 576 & 120 & 79 & 85.9 \\
            SparseVLM            & 64  & 81 & 48 & 75.1 \\
            VisionZip            & 64  & 77 & 39 & 77.0 \\
            PruneSID             & 64  & 78 & 41 & 84.1 \\
            \textbf{ASAP (Ours)} & \textbf{64} & \textbf{64} & \textbf{35} & \textbf{85.7} \\
            \bottomrule
        \end{tabular}
        }
    \end{minipage}
\end{table}

\subsection{Video Understanding}\label{subsec:video}

\begin{figure}[t]
    \centering
    \includegraphics[width=\textwidth]{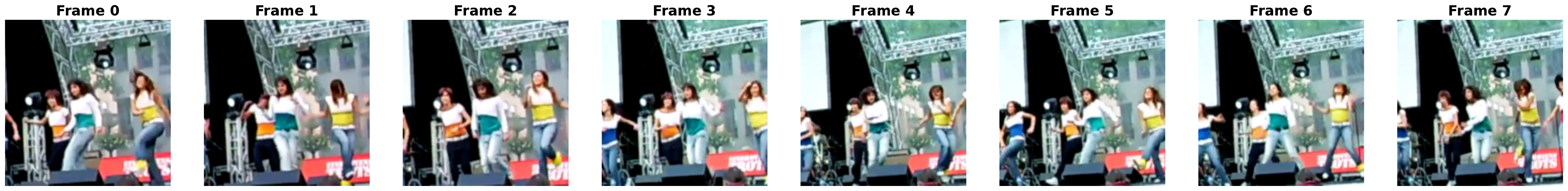} 
    \includegraphics[width=\textwidth]{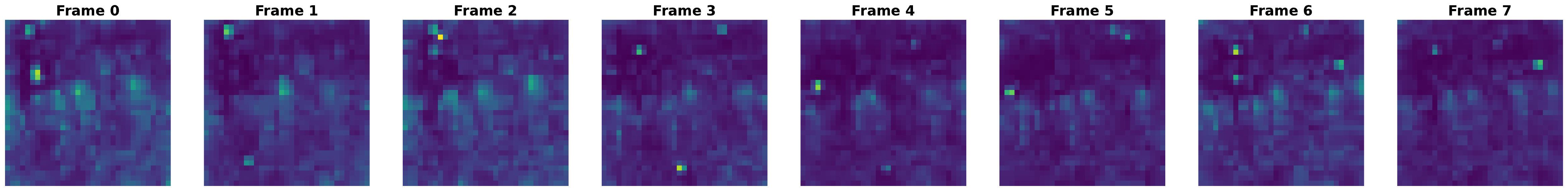}
    \includegraphics[width=\textwidth]{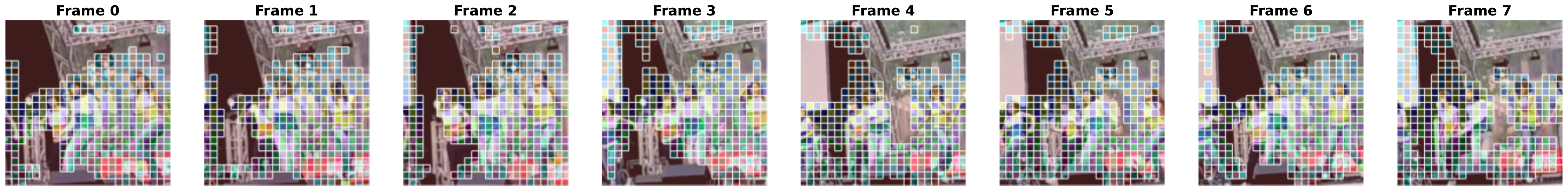}
    \caption{Qualitative results on Kinetics-400 using CLIP ViT. (Top) Original sequence. (Middle) The attention sink (yellow) shifts dynamically across frames. (Bottom) Our method tracks this shift and recomputes clustering per frame, consistently preserving moving foreground objects.}
    \label{fig:video_overtime}
    % \vspace{-4mm}
\end{figure}

To assess whether our method generalizes beyond classification-specific methods, we apply it to CLIP ViT~\cite{radford2021learning}, a feature extractor without a classification head, as used in LLaVA~\cite{liu2024improved}. For qualitative evaluation, we use the pretrained CLIP ViT directly. 
For quantitative evaluation, we freeze the backbone, attach a lightweight classification head, and fine-tune it for 10 epochs on Kinetics-400~\cite{kay2017kinetics}. The token reduction process remains strictly training-free; the linear head is fine-tuned solely for evaluation purposes, applied identically across all benchmarked methods.

As shown in Figure~\ref{fig:video_overtime}, despite CLIP ViT not being trained for classification, attention sinks emerge at semantically irrelevant background regions. Notably, the sink location shifts dynamically as the scene changes. Our method adapts accordingly by recomputing the diffusion-distance-based clustering per frame, ensuring continuous foreground preservation. Quantitatively, our method achieves only a $-0.21\%$ accuracy drop on Kinetics-400 (Table~\ref{tab:video-results}), outperforming baselines at comparable compression ratios. This confirms applicability to diverse vision encoders.

\subsection{Vision Language Model Experiments}\label{subsec:VLM}

Since VLMs share the CLIP ViT encoder analyzed in Section~\ref{subsec:video}, the foreground--background separation observed there directly transfers to multimodal reasoning. We apply our method to LLaVA~\cite{liu2024improved} and LLaVA-NeXT~\cite{liu2024llavanext}, and evaluate on up to seven image-based benchmarks: VQAv2~\cite{goyal2017making}, GQA~\cite{hudson2019gqa}, VizWiz~\cite{gurari2018vizwiz}, ScienceQA-IMG~\cite{lu2022learn}, POPE~\cite{li2023evaluating}, MME~\cite{fu2023mme}, and MMBench~\cite{liu2024mmbench}.

\begin{table}[t]
    \centering
    % \caption{Performance comparison of VLM token reduction methods on various benchmarks.}
    \caption{Comparison on LLaVA-1.5-7B}
    \label{tab:vlm_reduction_results}
    \resizebox{0.9\linewidth}{!}{
        \begin{tabular}{lccccccccccc}
            \toprule
            \multirow{2}{*}{Method} & \multirow{2}{*}{\begin{tabular}[c]{@{}c@{}}Remain \\ Tokens\end{tabular}} & \multicolumn{7}{c}{Benchmarks ($\uparrow$)} & \multirow{2}{*}{Avg.} & \multirow{2}{*}{Rel (\%)} \\
            \cmidrule(lr){3-9}
            & & GQA & MME & POPE & SQA-Img & VizWiz & VQAv2 & MMB-En & \\
            \midrule
            LLaVA (Full) & 576 & 61.9 & 1862 & 85.9 & 69.5 & 54.3 & 78.5 & 64.7 & 72.6 & 100 \\
            \midrule
            ToMe \citep{bolya2023token} & 128 & 52.4 & 1343 & 62.8 & 59.6 & 50.0 & 63.0 & 53.3 & 58.3 & 80.3 \\
            FastV \citep{chen2024image} & 128 & 49.6 & 1490 & 59.6 & 60.2 & 51.3 & 61.8 & 56.1 & 59.0 & 81.3 \\
            SparseVLM \citep{zhang2024sparsevlm} & 128 & 56.0 & 1696 & 80.5 & 67.1 & 51.4 & 73.8 & 60.0 & 67.7 & 93.3 \\
            VisionZip \citep{yang2025visionzip} & 128 & 57.6 & 1762 & 83.2 & 68.9 & 54.5 & \textbf{75.6} & 62.0 & 70.0 & 96.4 \\
            VisPruner \citep{zhang2025beyond} & 128 & 58.0 & 1771 & 84.6 & \textbf{69.1} & 52.7 & 73.9 & 61.9 & 69.8 & 96.1 \\
            PruneSID \citep{fang2026prune} & 128 & 58.9 & 1760 & \textbf{86.9} & 68.8 & 56.0 & 75.4 & 62.6 & 70.9 & 97.7 \\
            \textbf{ASAP (Ours - Pruning)} & 128 & 59.4 & 1784 & 86.0 & 68.8 & 56.3 & 74.9 & 61.9 & 70.9 & 97.7 \\
            \textbf{ASAP (Ours - Pooling)} & 128 & \textbf{60.4} & \textbf{1812} & 86.2 & \textbf{69.1} & \textbf{56.7} & 74.9 & \textbf{62.7} & \textbf{71.5} & \textbf{98.5} \\
            \midrule
            ToMe \citep{bolya2023token} & 64 & 48.6 & 922 & 52.5 & 50.0 & 50.2 & 57.1 & 43.7 & 49.8 & 68.6 \\
            FastV \citep{chen2024image} & 64 & 46.1 & 1020 & 48.0 & 51.1 & 50.8 & 55.0 & 48.0 & 50.0 & 68.9 \\
            SparseVLM \citep{zhang2024sparsevlm} & 64 & 52.7 & 1221 & 75.1 & 62.2 & 50.1 & 68.2 & 56.2 & 60.8 & 83.7 \\
            VisionZip \citep{yang2025visionzip} & 64 & 55.2 & 1694 & 77.1 & 69.2 & 54.9 & 72.8 & 60.1 & 67.7 & 93.2 \\
            VisPruner \citep{zhang2025beyond} & 64 & 55.4 & 1691 & 80.4 & \textbf{69.1} & 53.3 & 72.7 & 61.3 & 68.1 & 93.8 \\
            PruneSID \citep{fang2026prune} & 64 & 57.2 & 1734 & 84.1 & 68.1 & \textbf{57.0} & \textbf{73.8} & 59.7 & 69.5 & 95.7 \\
            % \textbf{ASAP (Ours)} & 64 & 58.1 & 1697 & 84.7 & 68.6 & 56.0 & 72.5 & 61.5 & 69.5 & 95.7 \\
            \textbf{ASAP (Ours) + Hybrid} & 64 & \textbf{59.6} & \textbf{1781} & \textbf{85.7} & 68.3 & 56.0 & 73.0 & \textbf{62.0} & \textbf{70.2} & \textbf{96.6} \\
            \bottomrule
        \end{tabular}
    }
    \vspace{-4mm}
\end{table}

Unlike image classification, the computational bottleneck in VLMs stems from both the vision encoder and the LLM decoder, as decoder cost scales linearly with the number of visual tokens. Our core pipeline compresses the background cluster $C_1$ via Transition Weight Pooling and samples the surviving foreground clusters ($C_2, \ldots, C_K$) at regular intervals in diffusion distance space, requiring no pairwise computation. At the 128-token budget, this core pipeline alone is sufficient. 

For extreme compression (64 tokens), ASAP first removes the geometrically identified background, after which existing reduction methods can be applied as a plug-in to the surviving foreground tokens. In our experiments, we apply bipartite matching using diffusion distance as the metric, pruning tokens whose information flow trajectories are most redundant with their matched pairs. A detailed component analysis is provided in Appendix~\ref{appendix:hybrid}.

As reported in Table~\ref{tab:vlm_reduction_results}, our pooling variant achieves 71.5\% average score at 128 tokens (98.5\% relative to full LLaVA), outperforming all compared methods in average score with the core pipeline alone. At 64 tokens, retaining only 11\% of the original visual tokens, the full pipeline achieves 70.2\% (96.6\% relative), with only a 1.3\%p gap from the 128-token result, while competing approaches degrade more. This validates that sink-anchored background removal efficiently allocates the limited token budget to meaningful foregrounds. 
Our POPE score at 128 tokens (86.2\%) exceeds the full baseline (85.9\%) and remains within 0.2\%p of it at 64 tokens (85.7\%); Section~\ref{sec:failure} analyzes where this gain arises and where fine-grained queries instead lose accuracy.
% POPE scores (86.2,85.7\% at 128/64 tokens) match or exceed the full baseline (85.9\%) suggesting background removal regularizes against LLM hallucination (Appendix~\ref{appendix:hallucination}).

\begin{table}[t]
    \centering
    \vspace{-10pt}
    \caption{Comparison on LLaVA-NeXT-7B.}
    \label{tab:llava_next}
    \resizebox{0.9\linewidth}{!}{
        \begin{tabular}{lcccccccccc}
            \toprule
            \multirow{2}{*}{Method} & \multirow{2}{*}{\begin{tabular}[c]{@{}c@{}}Remain \\ Tokens\end{tabular}} & \multicolumn{6}{c}{Benchmarks ($\uparrow$)} & \multirow{2}{*}{Avg.} & \multirow{2}{*}{Rel (\%)} \\
            \cmidrule(lr){3-8}
            & & GQA & MME & POPE & SQA-Img & VQAv2 & MMB-En & \\
            \midrule
            LLaVA-Next (Full) & 2880 & 64.2 & 1842 & 86.4 & 70.2 & 80.1 & 67.9 & 76.8 & 100 \\
            \midrule
            VisionZip \citep{yang2025visionzip} & 640 & 61.3 & 1787 & 86.3 & 68.1 & 79.1 & \textbf{66.3} & 75.1 & 97.8 \\
            PruneSID \citep{fang2026prune} & 640 & 61.6 & 1795 & 86.3 & \textbf{68.3} & 78.5 & 64.2 & 74.8 & 97.4 \\
            \textbf{ASAP (Ours)} & 640 & \textbf{62.3} & \textbf{1845} & \textbf{87.2} & 67.8 & \textbf{79.9} & 65.8 & \textbf{75.9} & \textbf{98.8} \\
            \midrule
            VisionZip \citep{yang2025visionzip} & 320 & 59.3 & 1702 & 82.1 & 67.3 & 76.2 & 63.1 & 72.2 & 94.0 \\
            PruneSID \citep{fang2026prune} & 320 & 60.5 & 1754 & 83.1 & 67.3 & 76.6 & 63.0 & 73.0 & 95.1 \\
            \textbf{ASAP (Ours) + Hybrid} & 320 & \textbf{61.0} & \textbf{1776} & \textbf{84.8} & \textbf{67.7} & \textbf{77.1} & \textbf{64.5} & \textbf{74.0} & \textbf{96.3} \\
            \midrule
            VisionZip \citep{yang2025visionzip} & 160 & 55.5 & 1630 & 74.8 & \textbf{68.3} & 71.4 & 60.1 & 68.6 & 89.3 \\
            PruneSID \citep{fang2026prune} & 160 & \textbf{58.9} & 1704 & 76.9 & 67.1 & 73.8 & 60.8 & 70.5 & 91.7 \\
            \textbf{ASAP (Ours) + Hybrid} & 160 & 57.0 & \textbf{1730} & \textbf{81.6} & 67.8 & \textbf{74.8} & \textbf{61.8} & \textbf{71.6} & \textbf{93.2} \\
            \bottomrule
        \end{tabular}
    }
    \vspace{-10pt}
\end{table}

These gains scale to higher resolutions: on LLaVA-NeXT (up to 2880 tokens, 5$\times$ that of LLaVA-1.5; Table~\ref{tab:llava_next}), ASAP outperforms VisionZip and PruneSID at 640, 320, and 160 token budgets. The MME score at 640 tokens (1845) again exceeds the full baseline (1842), and at 160 tokens (5.6\% retention) ASAP retains 93.2\% relative performance, with a particularly large POPE margin (81.6\% vs.\ 74.8\% for VisionZip).

\subsection{Hyperparameter Analysis}\label{subsec:hyperparmas}

\textbf{Hyperparameter Sensitivity.} ASAP introduces two primary hyperparameters: the cluster count $K$ and the sink detection threshold $\tau$. 
(We fix $\alpha = 0.5$ following the convention of Attention Rollout~\cite{abnar2020quantifying}; sensitivity analysis for $\alpha$ is provided in Appendix~\ref{appendix:alpha}.)
Figure~\ref{fig:ablation_pareto} shows the accuracy--FLOPs trade-off on DeiT-Base across $K \in \{3, \dots, 6\}$ and $\tau \in \{4,\dots,10\}$.

\begin{wrapfigure}{r}{0.35\textwidth}
    \vspace{-5pt}
    \centering
    \includegraphics[width=0.34\textwidth]{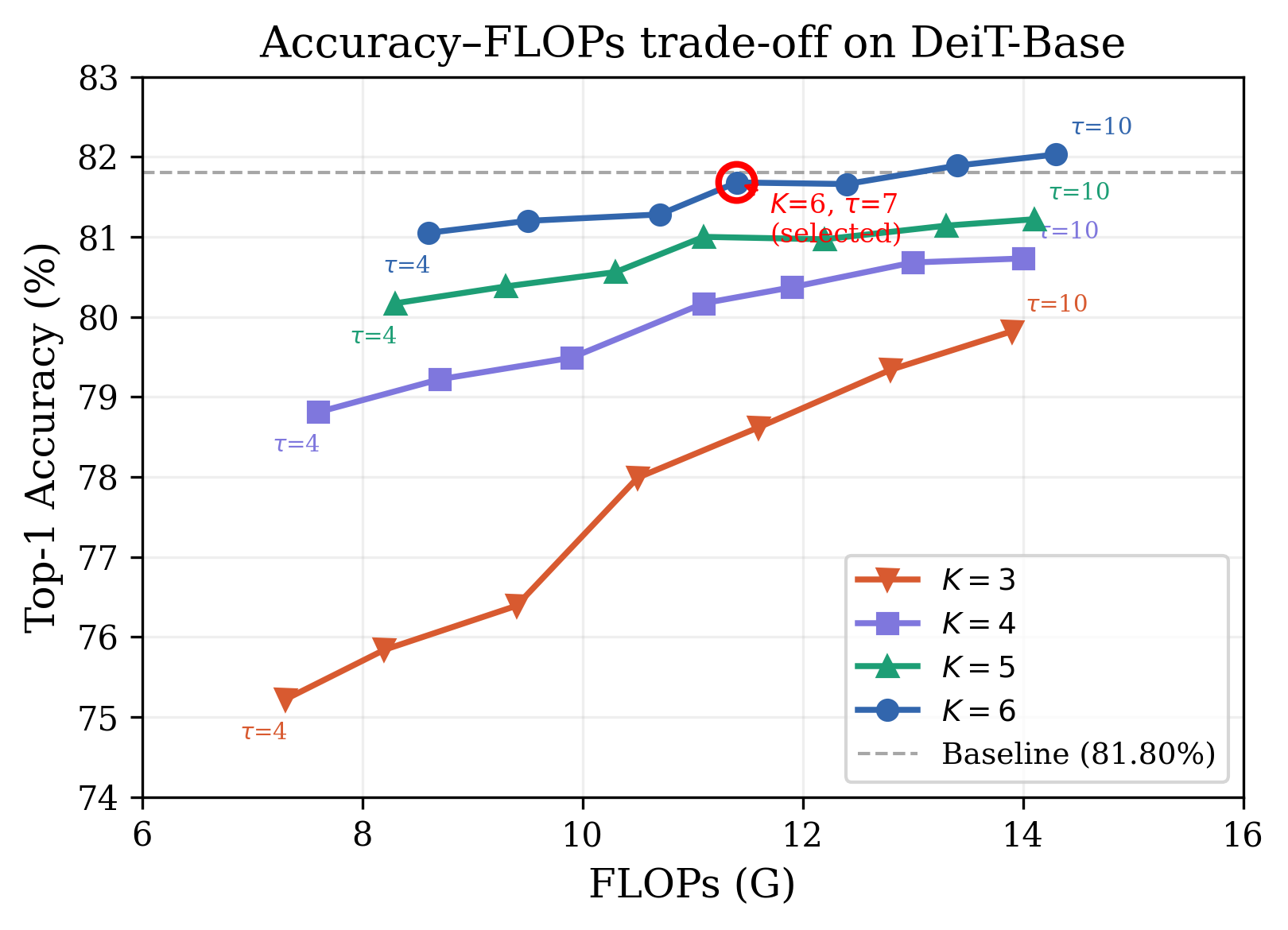}
     \caption{Accuracy--FLOPs trade-off on DeiT-Base for varying $K$ and $\tau$. The red circle marks the selected operating point ($K{=}6$, $\tau{=}7$).}
    \label{fig:ablation_pareto}
    \vspace{-15pt}
\end{wrapfigure}

\textbf{$K$ admits a clear lower bound}: $C_1$ captures foreground at $K{=}3$ (79.82\%) and vanishes as $K \to N$; the best point ($K{=}6$, $\tau{=}10$) reaches 82.03\%, slightly exceeding baseline.

\textbf{Robustness.}
Within $K \in \{5,6\}$ and $\tau \in \{7,8,9\}$, the total accuracy variation is only 0.92\%p. We select $K{=}6$, $\tau{=}7$ as the default (red circle in Figure~\ref{fig:ablation_pareto}), balancing accuracy retention ($-0.12$\%) with 35.2\% FLOPs reduction.

\textbf{Role of $\tau$.}
For each $K$, increasing $\tau$ monotonically raises both accuracy and FLOPs, confirming $\tau$ controls the accuracy--efficiency trade-off: a larger $\tau$ delays sink detection to deeper layers, 
allowing %the cumulative transition matrix 
$P^{(t^*)}$ to encode richer global context at the cost of additional computation. We observe the same trend on ViT-AugReg, where $K$'s lower bound is consistently reproduced (see Appendix~\ref{appendix:augreg_ablation}).

\subsection{Efficiency Analysis}\label{sec:latency}

Existing methods such as VisionZip~\cite{yang2025visionzip} delay token compression until the penultimate encoder layer, forcing the full quadratic attention cost through most of the network. In contrast, ASAP detects the attention sink at $t^*$ (typically layers 12--14 for CLIP ViT-Large) and compresses tokens early, reducing the sequence length for all subsequent layers and the LLM decoder. 
% Importantly, the practical bottleneck in VLMs is not the one-time transition matrix accumulation, but the repeated quadratic attention cost across deep encoder layers and the autoregressive decoder whose cost scales linearly with visual token count. 
Importantly, the bottleneck in VLMs is not the one-time accumulation but the repeated quadratic attention cost across encoder layers and the linearly-scaling decoder.
To quantify this structural advantage, we measure end-to-end inference latency on the POPE benchmark using a single NVIDIA A100-80GB GPU. As reported in Table~\ref{tab:latency_comparison}, ASAP achieves the fastest average inference time (64ms) and prefill time (35ms) under a 64-token budget, reducing prefill latency by 10\% compared to VisionZip while retaining substantially higher accuracy. 

As confirmed by our LLaVA-NeXT experiments (Appendix~\ref{appendix:latency}), this early-compression advantage widens as the input token count grows from 576 to 2880, with ASAP's prefill speedup over PruneSID increasing from 17\% to 35\%. This shows that the downstream savings from early token compression increasingly outweigh the fixed accumulation cost. %A general overhead analysis is provided in Appendix~\ref{appendix:overhead_breakdown}, and a detailed modular computational cost breakdown is presented in Appendix~\ref{appendix:cca}.
Detailed overhead and modular cost analyses are in Appendices~\ref{appendix:overhead_breakdown} and \ref{appendix:cca}.

\subsection{Failure Analysis}
\label{sec:failure}

Since ASAP compresses the region around the sink, we examine where this design costs accuracy, in both classification and VLM reasoning.

\paragraph{Scarce Background.}
We split ImageNet-val into deciles of a model-independent background ratio ($1 -$ union of ground-truth boxes / image area) and compare full DeiT-Base and ASAP within a single evaluation pipeline (Appendix~\ref{app:failure}).
The accuracy drop grows as the background shrinks, reaching $-0.56$\%p in the most foreground-dominant decile versus $-0.14$\%p overall, where the sink necessarily forms on the object and $C_1$ absorbs a homogeneous part of it.
The drop remains small because $C_1$ is pooled rather than discarded.
In the most background-rich deciles, ASAP matches or slightly exceeds the full model ($+0.30$ / $+0.48$\%p in deciles 8 / 10). Overall, only 0.81\% of validation images flip from correct to incorrect under ASAP.

\paragraph{Fine-Grained Details in VLMs.}
The same mechanism has two opposite effects in VLMs.
On POPE, the gain over the full LLaVA-1.5 grows with negative-sampling difficulty and concentrates on the adversarial split ($+0.01 \rightarrow +0.13 \rightarrow +0.96$ F1 for random / popular / adversarial), where co-occurrence-driven false positives are most frequent.
In contrast, a paired comparison on a fixed 10k-question VQAv2 subset (128 tokens) shows an overall drop of $-2.22$ points (95\% CI $[-2.7, -1.7]$), concentrated on questions that require fine local detail: text/OCR ($-6.1$) and color ($-5.7$), versus $-1.3$ to $-1.8$ for existence, counting, and spatial questions.
Pooling thus preserves scene-level and object-presence information but can lose small high-frequency details such as text, a limitation shared by token compression methods in general (Appendix~\ref{app:hallu-quant}).

\section{Discussion}

Beyond efficiency gains, background compression is associated with fewer hallucinated answers on the evaluated benchmarks (Section~\ref{sec:failure}). The cluster count $K$ admits natural bounds (Section~\ref{subsec:hyperparmas}): too few clusters cause $C_1$ to absorb foreground tokens ($\approx$2\%p drop at $K{=}3$), while sufficiently many clusters can exceed baseline accuracy on both DeiT-Base and ViT-AugReg (Appendix~\ref{appendix:augreg_ablation}). A preliminary Otsu-based study shows that $K$ can be derived per image from the distance distribution (Appendix~\ref{app:adaptivek}); a comparison at matched budgets remains future work.

% \paragraph{Background Compression as Visual Grounding.}
\textbf{Background Compression as Visual Grounding.}
An unexpected finding is that geometric background removal not only preserves accuracy but can improve it on hallucination-sensitive benchmarks: on POPE, ASAP exceeds the full-token baseline at 128 tokens (86.2\% vs.\ 85.9\%) and nearly matches it at 64 tokens (85.7\%). We hypothesize two mechanisms: attention sinks aggregate diffuse global features with minimal local information~\cite{darcet2024vision}, injecting visual noise into the LLM decoder; and compressing $C_1$ into a single token yields a sparser, higher-signal context. 
Together with the corrected cases and the split-wise POPE analysis (Appendix~\ref{appendix:hallucination}), this suggests a grounding benefit for object-presence queries, while fine-grained details remain the main cost of compression (Section~\ref{sec:failure}).

% \paragraph{Artifacts as Structural Signals.}
\textbf{Artifacts as Structural Signals.}
More broadly, model artifacts that emerge consistently across architectures need not be treated as pathologies. The attention sink---previously studied either as noise to mitigate~\cite{xiao2023efficient} or a behavior to replace via added registers~\cite{darcet2024vision}---here functions as a coordinate origin for the diffusion geometry of $P^{(t^\ast)}$. The predictability of its emergence is the signal we exploit.

\textbf{Limitations.}
Our method assumes that the dominant sink forms in a semantically neutral background region, which~\cite{darcet2024vision} reports to be the predominant case across architectures, with high-norm outlier tokens concentrating at low-informative patches near image borders.
Section~\ref{sec:failure} quantifies the cost when this assumption weakens: the accuracy drop grows in foreground-dominant images, where $C_1$ absorbs a homogeneous part of the object itself, and fine-grained details such as text can be lost in VLMs, although pooling $C_1$ keeps the classification drop small.
While our threshold-based detection always yields a well-defined sink, this anchor may be less informative in scenes without a dominant background.
Making $K$ adaptive, replacing $\tau$ with a quantile-based criterion, and extending to hybrid or non-patch-based encoders remain open.

\section{Conclusion}

We proposed ASAP, a training-free token reduction framework that models the attention sink as a mass-accumulating attractor in a Markov chain formulation and exploits it as the geometric anchor for token clustering. 
% By computing diffusion distances within the cumulative transition matrix, ASAP achieves foreground-background separation in a single forward pass at $t^\ast$---without requiring any fine-tuning or heuristic layer selection---and consistently outperforms existing training-free methods across image classification, video understanding, and vision-language models. 
By computing diffusion distances within the cumulative transition matrix, ASAP achieves foreground-background separation in a single forward pass at $t^{*}$. It achieves the best average accuracy retention among training-free methods across diverse visual modalities, without fine-tuning the backbone or manually selecting the reduction layer.
Notably, at 22\% of the original tokens, our method retains 98.5\% of the full VLM baseline and exceeds it on hallucination-sensitive benchmarks, demonstrating that what prior work treated as a model artifact can be leveraged as an organizing structure for efficient token reduction. Future directions include deriving the cluster count $K$ adaptively from the diffusion distance distribution and extending the Markov-chain-based framework to KV cache compression in large language models.

\section*{Acknowledgment}
This work was supported by the National Research Foundation of Korea (NRF) grant funded by the Korea government (MSIT) (RS-2025-24873052).

\bibliography{aimlk}
\bibliographystyle{unsrt}

% References follow the acknowledgments in the camera-ready paper. Use unnumbered first-level heading for
% the references. Any choice of citation style is acceptable as long as you are
% consistent. It is permissible to reduce the font size to \verb+small+ (9 point)
% when listing the references.
% Note that the Reference section does not count towards the page limit.
% \medskip

% {
% \small

% [1] Alexander, J.A.\ \& Mozer, M.C.\ (1995) Template-based algorithms for
% connectionist rule extraction. In G.\ Tesauro, D.S.\ Touretzky and T.K.\ Leen
% (eds.), {\it Advances in Neural Information Processing Systems 7},
% pp.\ 609--616. Cambridge, MA: MIT Press.

% [2] Bower, J.M.\ \& Beeman, D.\ (1995) {\it The Book of GENESIS: Exploring
%   Realistic Neural Models with the GEneral NEural SImulation System.}  New York:
% TELOS/Springer--Verlag.

% [3] Hasselmo, M.E., Schnell, E.\ \& Barkai, E.\ (1995) Dynamics of learning and
% recall at excitatory recurrent synapses and cholinergic modulation in rat
% hippocampal region CA3. {\it Journal of Neuroscience} {\bf 15}(7):5249-5262.
% }

%%%%%%%%%%%%%%%%%%%%%%%%%%%%%%%%%%%%%%%%%%%%%%%%%%%%%%%%%%%%

\newpage
\appendix

\section{Preliminaries on Diffusion Distance}\label{appendix:diffusion_prelim}
 
We provide a brief review of diffusion distance~\cite{coifman2006diffusion} to make the paper self-contained. Given a Markov chain with transition matrix $P$, the diffusion distance between two states $x_i$ and $x_j$ at time $t$ is defined as:
\begin{equation}\label{eq:weighted_diff}
    D_t(x_i, x_j)^2 = \sum_k \frac{\left(P^{(t)}_{i,k} - P^{(t)}_{j,k}\right)^2}{\phi(k)}
\end{equation}
where $\phi(k)$ is the stationary distribution serving as a weighting over states. Intuitively, $D_t(x_i, x_j)$ measures how differently two states distribute their probability mass after $t$ steps of the random walk. Two states are close in diffusion distance if and only if they route information to the same destinations with similar probabilities---capturing connectivity through all possible multi-hop paths on the graph, rather than direct pairwise similarity alone.
 
A key result of~\cite{coifman2006diffusion} is that under a uniform stationary measure, this reduces to the unweighted $\ell_2$ distance between probability rows:
\begin{equation}\label{eq:unweighted_diff}
    D_t(x_i, x_j) = \left\| P^{(t)}_{i,*} - P^{(t)}_{j,*} \right\|_2
\end{equation}
which bypasses the spectral decomposition originally required and enables efficient computation. In our framework, we apply this distance specifically between each token $x_i$ and the attention sink $x_s$ (Eq.~\ref{eq:diffusion_dist}), reducing the problem from $O(N^2d)$ pairwise comparisons to $N$ scalar distances from a single reference point.
 
\paragraph{Practical justification of the uniform measure approximation.}

The exact diffusion distance (Eq.~\ref{eq:weighted_diff}) requires the stationary distribution $\phi(k)$, whose computation involves eigendecomposition at $O(N^3)$ cost---defeating the efficiency goal of token pruning. We adopt the unweighted approximation (Eq.~\ref{eq:unweighted_diff}) based on both theoretical and empirical justification.

Theoretically, ASAP computes distances only from a single reference point (the sink), not all $O(N^2)$ pairs. The reweighting factor $1/\phi(k)$ in Eq.~\ref{eq:weighted_diff} primarily rescales dimensions corresponding to high-stationary-mass tokens (i.e., the sink itself), but since the sink serves as a fixed anchor rather than a comparison target among peers, this rescaling has limited effect on the relative ordering of token-to-sink distances. Furthermore, Radial Diffusion Clustering (Eq.~\ref{eq:clustering}) depends only on the \textit{rank ordering} of distances, not their absolute magnitudes, making it inherently robust to monotonic distortions.

Empirically, we verify this by computing the exact weighted diffusion distance using the column-mean approximation of $\phi(k)$ on 1,000 randomly sampled ImageNet validation images (DeiT-Base, $\tau=7$). 
The Spearman rank correlation between unweighted and weighted distances is $\rho = 0.987 \pm 0.012$, confirming near-perfect rank preservation. This means that the cluster assignments produced by Eq.~\ref{eq:clustering} are virtually identical under both formulations, validating the uniform measure approximation as a practical and theoretically grounded choice for our framework.

\section{Conditions for Monotone Mass Accumulation at Sink Tokens}\label{appendix:proof1}
To formalize the intuition behind sink formation, we state a sufficient condition under an idealized attention assumption. We refer to such a token as a \emph{mass-accumulating attractor}: a relaxed analog of an absorbing state in a finite Markov chain, where instead of strict absorption ($P_{s,s}=1$), the column sum $C^{(t)}_s = \sum_{i=1}^N P^{(t)}_{i,s}$ increases monotonically with depth.

% \begin{proposition}[Sufficient Condition for Monotone Mass Accumulation]\label{appendix:propo1}
\paragraph{Theoretical Motivation: Sufficient Condition for Monotone Mass Accumulation}\label{appendix:propo1}
Let $A^l$ be the attention matrix at layer $l$, and let token $s$ satisfy the following empirically motivated condition: it consistently receives above-average attention from all other tokens: $A^l_{k,s} \geq \frac{1}{N} + \epsilon$ for all $k$, with some $\epsilon > 0$. Then, under the Lazy Random Walk formulation $\tilde{A}^l = \alpha A^l + (1-\alpha)I$, the total probability mass (column sum) $C^{(t)}_s = \sum_{i=1}^N P^{(t)}_{i,s}$ is monotonically non-decreasing as long as $C^{(t)}_s < 1 + N\epsilon$. That is, such a token structurally acts as a \emph{dominant accumulator} in the cumulative transition matrix $P^{(t)}$, progressively absorbing probability mass from all other tokens.
% \end{proposition}

% \begin{proof}
\paragraph{Justification}
In our Markov chain model, the probability mass at token $s$ after $t+1$ layers is updated by:
\begin{equation}
    C^{(t+1)}_s = \sum_{k=1}^N C^{(t)}_k \tilde{A}^{t+1}_{k,s}
\end{equation}
Substituting our Lazy Random Walk definition $\tilde{A}^{t+1}_{k,s} = \alpha A^{t+1}_{k,s} + (1-\alpha)\delta_{ks}$, we obtain:
\begin{equation}
    C^{(t+1)}_s = (1-\alpha)C^{(t)}_s + \alpha \sum_{k=1}^N C^{(t)}_k A^{t+1}_{k,s}
\end{equation}
The change in probability mass at layer $t+1$ is thus given by $\Delta C_s = \alpha \left( \sum_{k=1}^N C^{(t)}_k A^{t+1}_{k,s} - C^{(t)}_s \right)$.
Given that $\sum_{k=1}^N C^{(t)}_k = N$ (since each row of $P^{(t)}$ sums to 1) and the sink assumption $A^{t+1}_{k,s} \geq \frac{1}{N} + \epsilon$, we have:
\begin{equation}
    \sum_{k=1}^N C^{(t)}_k A^{t+1}_{k,s} \geq \left(\frac{1}{N} + \epsilon\right) \sum_{k=1}^N C^{(t)}_k = 1 + N\epsilon
\end{equation}
Therefore, as long as $C^{(t)}_s < 1 + N\epsilon$, the net probability influx $\Delta C_s$ remains strictly positive. Consequently, the column sum accumulates mass monotonically, and $s$ dominates the global information flow, structurally acting as a dominant accumulator in the cumulative transition matrix.
% \end{proof}

This characterization provides a sufficient condition under which mass-accumulating attractors form under the Markov chain formulation: once a token consistently receives diffuse, above-average attention, it is expected to accumulate dominant probability mass. In practice, we detect this convergence event using a column-sum threshold $\tau$ (Section~\ref{sec:sink}); the proposition provides a formal basis for the threshold-based detection used in our framework.
We note that the condition $A^l_{k,s} \geq \frac{1}{N} + \epsilon$ for all $k$ and $l$ is a sufficient condition 
that may not hold strictly in early layers, where attention patterns are near-uniform or locally biased. In practice, 
the sink accumulates dominant probability mass even under weaker conditions, as empirically confirmed by the phase 
transition observed in Figure~\ref{fig:sink_emergence}. The proposition thus provides a mechanistic explanation 
for the observed mass accumulation behavior rather than a tight characterization of the necessary conditions.
 
\section{Diffusion Distance and Trajectory Separation}\label{appendix:propo2}
\paragraph{Diffusion Distance and Trajectory Separation}
Let $s$ be a mass-accumulating attractor of the Markov chain defined by the cumulative transition matrix $P^{(t)}$. For any token $x_i$, the diffusion distance $D(x_i, x_s) = \|P^{(t^*)}_{i,*} - P^{(t^*)}_{s,*}\|_2$ satisfies the following properties:
\begin{enumerate}
    \item \textbf{Full Convergence:} $D(x_i, x_s) = 0 
    \iff x_i$ has been fully absorbed into the sink, yielding identical information routing 
    $P^{(t^*)}_{i,*} = P^{(t^*)}_{s,*}$.
    \item \textbf{Trajectory Separation:} Tokens with large $D(x_i, x_s)$ route information differently from the sink, as reflected by their geometric separation in the $P^{(t^*)}$ manifold.
\end{enumerate}
% \end{proposition}
 
\paragraph{Justification}
(1) Follows directly from the positive definiteness of the $\ell_2$ norm: $\|v\|_2 = 0 \iff v = \mathbf{0}$.
 
(2) A large separation $D(x_i, s) = \delta > 0$ implies $\|P^{(t^*)}_{i,*} - P^{(t^*)}_{s,*}\|_2 = \delta$, meaning there exists at least one token $j$ such that $|P^{(t^*)}_{i,j} - P^{(t^*)}_{s,j}| > 0$ --- i.e., $x_i$ 
routes non-negligible information to destinations that the sink does not reach, indicating a distinct information trajectory.
% \end{proof}

\section{Implementation Details}\label{appendix:details}

This section describes the implementation details of ASAP. The full pipeline comprises two stages. Algorithm~\ref{alg:asap_core} is the core pipeline that performs diffusion-distance-based Radial Clustering and $C_1$ Transition Weight Pooling, while Algorithm~\ref{alg:asap_hybrid} is a plug-in pruning stage applied when extreme compression ratios are required (e.g., 576 $\to$ 64 tokens). Importantly, Algorithm~\ref{alg:asap_hybrid} does not perform additional attention computation at a separate layer; it is applied immediately after Algorithm~\ref{alg:asap_core} removes $C_1$ at the same layer $t^*$, reusing the already-computed cumulative transition matrix $P^{(t^*)}$ and the surviving foreground tokens. That is, both stages execute sequentially at a single time step $t^*$, and all subsequent encoder layers process only the final compressed token set.

\begin{algorithm}[h]
\caption{ASAP: Diffusion Distance Clustering and $C_1$ Pooling}
\label{alg:asap_core}
\begin{algorithmic}[1]
\REQUIRE Patch tokens $X = [x_{cls}, x_1, \ldots, x_N]$, cumulative transition matrix $P^{(t^*)}$, cluster count $K$, background cluster count $p$, token budget $T$
\ENSURE Reduced token set $X'$ with $|X'| \leq T + 2$
\STATE \textcolor{gray}{\textit{\% Stage 1-A: Sink identification}}
\STATE $s \leftarrow \arg\max_{j \geq 1} P^{(t^*)}_{0, j}$ \COMMENT{Sink via [CLS] row}
\STATE \textcolor{gray}{\textit{\% Stage 1-B: Diffusion distance}}
\FOR{$i = 1, \ldots, N$}
    \STATE $D(x_i, x_s) \leftarrow \| P^{(t^*)}_{i, *} - P^{(t^*)}_{s, *} \|_2$
\ENDFOR
\STATE \textcolor{gray}{\textit{\% Stage 1-C: Radial Diffusion Clustering}}
% \STATE $\tilde{D}_i \leftarrow (D_i - D_{\min}) / (D_{\max} - D_{\min})$ \COMMENT{Normalize to $[0, 1]$}
% \STATE $c_i \leftarrow \min(\lfloor \tilde{D}_i \cdot K \rfloor, K-1)$ \COMMENT{Cluster id $\in \{0, \ldots, K-1\}$}
% --- Algorithm 1 line 8-9 수정 ---
\STATE $\tilde{D}_i \leftarrow (D_i - D_{\min}) / (D_{\max} - D_{\min} + \epsilon)$ \COMMENT{Normalize to $[0, 1]$, $\epsilon = 10^{-6}$}
\STATE $c_i \leftarrow \mathrm{clamp}(\lfloor \tilde{D}_i \cdot K \rfloor, 0, K-1)$ \COMMENT{Cluster id $\in \{0, \ldots, K{-}1\}$}
\STATE \textcolor{gray}{\textit{\% Stage 1-D: $C_1$ Transition Weight Pooling}}
\STATE $\mathcal{C}_{bg} \leftarrow \{x_i \mid c_i < p\}$; \quad $\mathcal{C}_k \leftarrow \{x_i \mid c_i = k\}$ for $k = p, \ldots, K-1$
\STATE $w_i \leftarrow \exp(D_i) / \sum_{x_j \in \mathcal{C}_{bg}} \exp(D_j)$ \COMMENT{Softmax over bg}
\STATE $\mathbf{m_{bg}} \leftarrow \sum_{x_i \in \mathcal{C}_{bg}} w_i \mathbf{f}_i$ \COMMENT{Pooled bg token}
\STATE \textcolor{gray}{\textit{\% Stage 1-E: Greedy stride sampling from sink-proximal to sink-distal clusters (Optional, only if $|\mathcal{C}_{fg}| > T$)}}
\STATE $\mathcal{S} \leftarrow \emptyset$ \COMMENT{Surviving foreground tokens}
\STATE $\mathcal{C}_{fg} \leftarrow \bigcup_{k=p}^{K-1} \mathcal{C}_k$
\IF{$|\mathcal{C}_{fg}| > T$}
    \FOR{$k = p, p+1, \ldots, K-1$}% \COMMENT{Sink-proximal first}
        \STATE Sort tokens in $\mathcal{C}_k$ by $\tilde{D}_i$ ascending
        \STATE $\text{stride}_k \leftarrow \lfloor |\mathcal{C}_k| / \lfloor T \cdot |\mathcal{C}_k| / |\mathcal{C}_{fg}| \rfloor \rfloor$
        \IF{$\text{stride}_k = 0$}
            \STATE $\text{stride}_k \leftarrow 1$ \COMMENT{Fallback: retain all tokens in this cluster}
        \ENDIF
        \STATE $\mathcal{S} \leftarrow \mathcal{S} \cup \mathcal{C}_k[0 :: \text{stride}_k]$
        \IF{$|\mathcal{S}| \geq T$}
            \STATE $\mathcal{S} \leftarrow \mathcal{S} \cup \bigcup_{j=k+1}^{K-1} \mathcal{C}_j$ \COMMENT{Fully retain remaining (more sink-distal) clusters}
            \STATE \textbf{break}
        \ENDIF
    \ENDFOR
\ELSE
    \STATE $\mathcal{S} \leftarrow \mathcal{C}_{fg}$
\ENDIF
\STATE \textcolor{gray}{\textit{\% Stage 1-F: Assembly}}
\STATE $X' \leftarrow [x_{cls}] \,\|\, \mathcal{S} \,\|\, [\mathbf{m_bg}]$
\STATE \RETURN $X'$
\end{algorithmic}
\end{algorithm}

\paragraph{Core pipeline (Algorithm~\ref{alg:asap_core}).}
Given the cumulative transition matrix $P^{(t^*)}$ accumulated via the Lazy Random Walk, we first identify the sink index $s$ by taking the argmax of the \texttt{[CLS]} token's attention row (Eq.~\ref{eq:sink_detection}). We then compute the diffusion distance $D(x_i, x_s) = \|P^{(t^*)}_{i,*} - P^{(t^*)}_{s,*}\|_2$ from each patch token $x_i$ to the sink (Eq.~\ref{eq:diffusion_dist}). The distances are normalized to $[0, 1]$ and scaled by $K$ to assign each token to one of $K$ semantic level sets (Eq.~\ref{eq:clustering}). The $p$ clusters closest to the sink are designated as the background set $\mathcal{C}_{\mathrm{bg}}$, whose tokens are compressed into a single representative token $\mathbf{m}_{\mathrm{bg}}$ via softmax-weighted pooling based on diffusion distance (Eq.~\ref{eq:twp}). The final output is the concatenation of the \texttt{[CLS]} token, the foreground cluster tokens, and the pooled background token.

At the 128-token budget, Algorithm~\ref{alg:asap_core} alone achieves sufficient performance (see Table \ref{tab:vlm_reduction_results}). When the number of foreground tokens exceeds the budget $T$ after $C_1$ pooling, we perform stride sampling within each foreground cluster $C_k$: tokens are sorted by diffusion distance and sampled at budget-proportional intervals (Algorithm~\ref{alg:asap_core}, Stage~5). This selects tokens at uniform spacing in diffusion distance space within each cluster, preserving semantic coverage while meeting the target budget. When only $C_1$ pooling is applied without stride sampling, an average of approximately 230 tokens are retained, achieving 99.5\% relative performance compared to full LLaVA (Table~\ref{tab:vlm_ablation}).

\paragraph{Plug-in pruning (Algorithm~\ref{alg:asap_hybrid}).}
When extreme compression is required (e.g., 64 tokens), a two-step plug-in pruning is applied immediately after $C_1$ removal in Algorithm~\ref{alg:asap_core}---within the same layer $t^*$, without any additional forward pass---by reusing the surviving foreground tokens $\mathcal{C}_{\mathrm{fg}}$ and the already-computed cumulative transition matrix $P^{(t^*)}$. This stage is a modular plug-in independent of ASAP's core contribution, and can be replaced with any existing token reduction method.

\begin{algorithm}[H]
\caption{ASAP Hybrid: Stage 2 Plug-in Pruning (after $C_1$ Removal)}
\label{alg:asap_hybrid}
\begin{algorithmic}[1]
\REQUIRE Foreground tokens $\mathcal{C}_{\mathrm{fg}}$ from Algorithm~\ref{alg:asap_core}, cumulative transition matrix $P^{(t^*)}$, target token count $T$, removal batch size $r$
\ENSURE Final token set $\mathbf{X}'$ with $|\mathbf{X}'| \leq T + 1$
\STATE $\mathcal{S} \gets \mathcal{C}_{\mathrm{fg}}$ \COMMENT{Initialize survivor set}
\STATE \textcolor{gray}{\textit{\% Stage 2-A: CLS attention top-$k$ filtering}}
\IF{$|\mathcal{S}| > 3T$}
    \FOR{each $x_i \in \mathcal{S}$}
        \STATE $a_i \gets P^{(t^*)}_{0,\,i}$ \COMMENT{CLS-to-patch attention}
    \ENDFOR
    \STATE $\mathcal{S} \gets \mathrm{top\text{-}k}(\mathcal{S},\; 3T,\; \text{by } a_i)$ \COMMENT{Keep $3T$ highest-attention tokens}
\ENDIF
\STATE \textcolor{gray}{\textit{\% Stage 2-B: Iterative redundancy-based pruning via diffusion distance}}
\STATE $n_{\mathrm{remove}} \gets |\mathcal{S}| - T$
\WHILE{$n_{\mathrm{remove}} > 0$}
    \STATE Partition $\mathcal{S}$ into $\mathcal{G}_A$ (even-indexed) and $\mathcal{G}_B$ (odd-indexed)
    \STATE $M_{ab} \gets \| P^{(t^*)}_{a,*} - P^{(t^*)}_{b,*} \|_2$ for $a \in \mathcal{G}_A,\; b \in \mathcal{G}_B$ \COMMENT{Diffusion dist.\ matrix}
    \STATE $\mathrm{score}_b \gets \min_{a \in \mathcal{G}_A} M_{ab}$ for each $b \in \mathcal{G}_B$ \COMMENT{Closest match in $\mathcal{G}_A$}
    \STATE $r' \gets \min(r,\; n_{\mathrm{remove}})$ \COMMENT{Cap at remaining quota}
    \STATE $\mathcal{R} \gets \{b \in \mathcal{G}_B : \mathrm{score}_b \text{ in smallest } r'\}$ \COMMENT{Most redundant tokens}
    \STATE $\mathcal{S} \gets \mathcal{S} \setminus \mathcal{R}$
    \STATE $n_{\mathrm{remove}} \gets n_{\mathrm{remove}} - r'$
\ENDWHILE
\STATE \textcolor{gray}{\textit{\% Assembly}}
\STATE $\mathbf{X}' \gets [\mathbf{x}_{\texttt{cls}}] \;\|\; \mathcal{S}$
\RETURN $\mathbf{X}'$
\end{algorithmic}
\end{algorithm}

In Stage~2-A, we select the top-$3T$ foreground tokens ranked by the \texttt{[CLS]} row of the cumulative transition matrix $P^{(t^*)}_{0,*}$. Since background tokens---including the attention sink---have already been removed in Stage~1, the \texttt{[CLS]} attention scores are extracted exclusively over foreground tokens, free from the distortion where the sink absorbs attention mass and causes foreground token importance to be underestimated. This enables reliable preservation of tokens with genuinely strong information connections to \texttt{[CLS]}. In Stage~2-B, the remaining tokens are partitioned into two groups $\mathcal{G}_A$ (even-indexed) and $\mathcal{G}_B$ (odd-indexed), and a diffusion distance matrix is computed between the two groups. Greedy bipartite matching then identifies the $n_{\mathrm{prune}}$ closest-matched pairs---tokens whose information flow trajectories are most redundant---and removes them to reach the target budget $T$.

This design offers two key advantages. First, since all reduction is completed at a single time step $t^*$, every encoder layer after $t^*\!+\!1$ and the entire LLM decoder process only the final compressed sequence, maximizing downstream computational savings---as demonstrated in Section~\ref{sec:latency} and Appendix~\ref{appendix:latency}. Second, as shown in the analysis of Appendix~\ref{appendix:hybrid}, replacing cosine similarity with diffusion distance in the plug-in stage improves relative performance from 95.4\% to 96.3\%, confirming that the global manifold information encoded in $P^{(t^*)}$ is consistently effective for redundancy identification as well.

\section{Failure Analysis on ImageNet}
\label{app:failure}

\begin{figure}[p]
  \centering
  {\small\textbf{(a) Background-rich images (decile 10): full model wrong, ASAP correct}}\\[3pt]
  \includegraphics[width=0.45\linewidth]{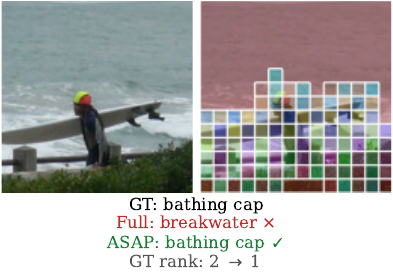}\hfill
  \includegraphics[width=0.45\linewidth]{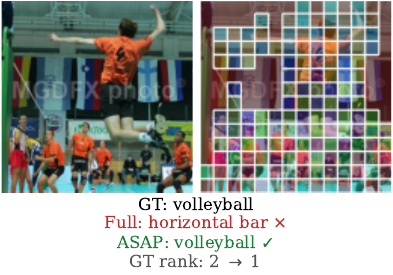}\\[3pt]
  \includegraphics[width=0.45\linewidth]{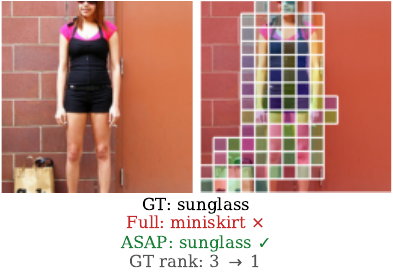}\hfill
  \includegraphics[width=0.45\linewidth]{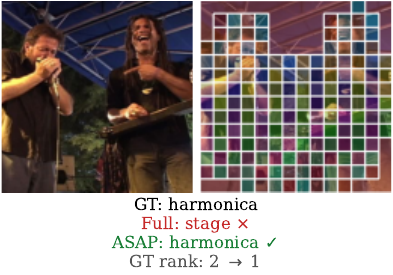}\\[10pt]
  {\small\textbf{(b) Foreground-dominant images (decile 1): full model correct, ASAP wrong}}\\[3pt]
  \includegraphics[width=0.45\linewidth]{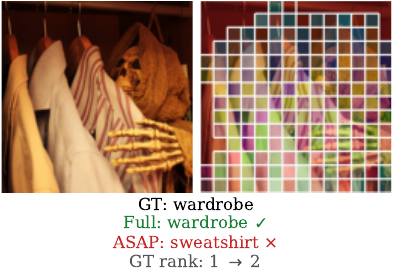}\hfill
  \includegraphics[width=0.45\linewidth]{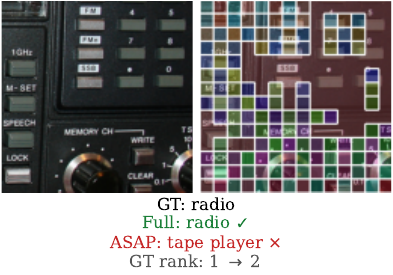}\\[3pt]
  \includegraphics[width=0.45\linewidth]{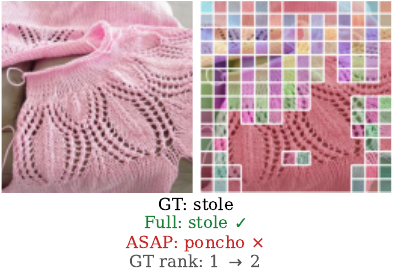}\hfill
  \includegraphics[width=0.45\linewidth]{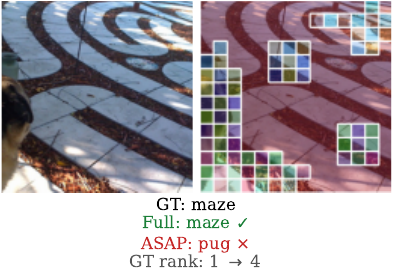}
  \caption{Examples from the two ends of the background-ratio spectrum (DeiT-Base). Each pair shows the input and the tokens ASAP retains (outlined; dimmed patches are pooled into $C_1$); GT rank is the rank of the ground-truth class in the top-5 of the full model $\rightarrow$ ASAP.
  (a) When the object is small, the full model predicts a class describing the surrounding scene (breakwater, horizontal bar, miniskirt, stage); after the background is pooled, ASAP predicts the ground-truth object.
  (b) When the object fills the image, there is no background to pool, so $C_1$ covers a large homogeneous part of the object itself (knitted fabric, the paving of the maze, the radio's front panel). The retained tokens still show local parts, and ASAP predicts a related class (poncho, sweatshirt, tape player), with the ground truth typically dropping to second place.}
  \label{fig:failure}
\end{figure}

\paragraph{Protocol.}
For each ImageNet-val image, we compute a model-independent background ratio, $1 - |\bigcup_b B_b| / |I|$, from the official localization boxes $\{B_b\}$, and split the 50{,}000 images into deciles (decile 1: least background).
The full DeiT-Base model and ASAP ($K{=}6$, $\tau{=}7$) are evaluated in a single pipeline.
In this pipeline the full-validation accuracies are 81.74\% and 81.60\% ($-0.14$\%p), within 0.1\%p of Table~\ref{tab:vit_results} (81.80\% / 81.68\%); all comparisons below use this pipeline.

\begin{table}[h]
\centering
\small
\setlength{\tabcolsep}{3.5pt}
\caption{Top-1 accuracy (\%) by background-ratio decile (decile 1: least background), with the discordant pairs and the two-sided exact McNemar $p$-value per decile.}
\label{tab:decile}
\begin{tabular}{lcccccccccc}
\toprule
 & 1 & 2 & 3 & 4 & 5 & 6 & 7 & 8 & 9 & 10 \\
\midrule
Full & 80.80 & 83.46 & 83.92 & 85.16 & 85.00 & 84.68 & 84.02 & 82.44 & 78.96 & 69.00 \\
ASAP & 80.24 & 83.10 & 83.56 & 85.00 & 84.88 & 84.24 & 83.96 & 82.74 & 78.82 & 69.48 \\
$\Delta$ & $-0.56$ & $-0.36$ & $-0.36$ & $-0.16$ & $-0.12$ & $-0.44$ & $-0.06$ & $+0.30$ & $-0.14$ & $+0.48$ \\
\midrule
Full only & 57 & 39 & 37 & 35 & 31 & 40 & 39 & 36 & 48 & 45 \\
ASAP only & 29 & 21 & 19 & 27 & 25 & 18 & 36 & 51 & 41 & 69 \\
McNemar $p$ & .003 & .027 & .022 & .37 & .50 & .005 & .82 & .13 & .53 & .031 \\
\bottomrule
\end{tabular}
\end{table}

% \begin{figure}[h] ... decile Δ 막대 그래프 + sink overlay (실패 2 / 유지 2 예시) ... \end{figure}

\paragraph{Results.}
From decile 4 to decile 1 the drop grows from $-0.16$ to $-0.56$\%p, four times the overall drop.
It stays small because $C_1$ is pooled into one token rather than discarded.
In the most background-rich deciles ASAP slightly exceeds the full model ($+0.30$ and $+0.48$\%p in deciles 8 and 10), nominally significant only in decile 10 (exact McNemar test on the discordant pairs, Table~\ref{tab:decile}: $p{=}0.13$ and $p{=}0.03$; neither survives a correction for testing ten deciles). The drop in the most foreground-dominant decile, in contrast, is significant ($p{=}0.003$; 57 images flip from correct to incorrect versus 29 in the opposite direction).
Decile 10, where small objects sit in large backgrounds, is also the hardest decile for the full model (69.00\%).
Across the validation set, 407 images (0.81\%) are classified correctly by the full model and incorrectly by ASAP.

\paragraph{Qualitative inspection.}
In the failure cases, the image contains almost no background, so $C_1$ covers a large homogeneous part of the object itself; the remaining local parts lead ASAP to a related class (Figure~\ref{fig:failure}b).
Foreground-dominant images that are still classified correctly typically have the sink inside a homogeneous region of the object (e.g., a uniformly colored surface). Figure~\ref{fig:failure} shows decile-1 failures (b) alongside decile-10 images where pooling the background corrects the full model (a).

\section{Computational Cost Analysis}\label{appendix:overhead}

\subsection{Accumulation and Clustering Overhead}\label{appendix:overhead_breakdown}

The computational cost of our method comprises two components: (i)~cumulative transition matrix accumulation (Eq.~\ref{eq:lazy_rw}), and (ii)~diffusion distance computation and Radial Diffusion Clustering (Eqs.~\ref{eq:diffusion_dist}--\ref{eq:clustering}).

\paragraph{Matrix accumulation cost.}
Computing $P^{(t^*)} = \tilde{A}^1 \times \cdots \times \tilde{A}^{t^*}$ requires $t^* - 1$ multiplications of $N \times N$ matrices, where $N$ is the number of patch tokens. For the standard ViT setting with $16 \times 16$ patches at $224 \times 224$ resolution, $N = 196$. Each $196 \times 196$ matrix multiplication requires approximately $2 \times 196^3 \approx 1.5 \times 10^7$ FLOPs --- roughly $13\%$ of a single self-attention layer ($\sim 4N^2 d \approx 1.2 \times 10^8$ FLOPs for $d = 768$, DeiT-Base), and a small fraction of the full forward pass since accumulation terminates at $t^*$, not at the final layer $L$. In our experiments, $t^*$ is typically $8$--$9$ for DeiT ($L = 12$) and $12$--$14$ for CLIP ViT-Large ($L = 24$), meaning only a fraction of layers contribute to the accumulation cost.

\paragraph{Clustering cost.}
Given $P^{(t^*)}$, computing $N$ diffusion distances (Eq.~\ref{eq:diffusion_dist}) requires $O(N^2)$ operations (one $\ell_2$ norm per token against the sink row). The subsequent Radial Diffusion Clustering (Eq.~\ref{eq:clustering}) involves sorting ($O(N \log N)$) and binning ($O(N)$). For $N = 196$, the total clustering time is $<$0.01\,ms.

\paragraph{Empirical measurement.}
To provide an upper bound on the accumulation overhead, Table~\ref{tab:cost} reports throughput with accumulation running across \textit{all} layers without early stopping. Under this worst-case setting, the additional FLOPs are negligible (DeiT-B: no measurable increase; CLIP ViT-L: $+0.1$G), while throughput decreases by $11\%$ (DeiT-B) and $7\%$ (CLIP ViT-L). These figures represent strict upper bounds; with early stopping at $t^*$, the actual overhead is proportionally smaller (roughly $t^*/L \approx 60\%$ of the reported values for DeiT).
\begin{wraptable}{r}{0.45\textwidth}
    % \vspace{-4mm} % 텍스트 위쪽 여백 조정 (필요시 수치 변경)
    \centering
    \caption{Computational Cost Analysis}
    \label{tab:cost}
    \resizebox{\linewidth}{!}{
        \begin{tabular}{lcc}
            \toprule
            Model & FLOPs (G) & Throughput (image/s) \\
            \midrule
            DeiT-B & 17.6 & 1206.13 \\
            + Lazy Random Walk & 17.6 & 1077.52 \\
            \midrule
            ClipViT-L & 174.6 & 110.09 \\
            + Lazy Random Walk & 174.7 & 102.38 \\
            \bottomrule
        \end{tabular}
    }
    \vspace{-1em} % 텍스트 아래쪽 여백 조정 (필요시 수치 변경)
\end{wraptable}
Importantly, this overhead is far outweighed by the throughput gains from token reduction. For DeiT-B, the end-to-end pipeline (accumulation $+$ clustering $+$ pruning) yields a net throughput improvement of $48\%$ (Table~\ref{tab:vit_results}), and for CLIP ViT-L on Kinetics-400, the net improvement is $24.5\%$ (Table~\ref{tab:video-results}).

\paragraph{Scaling considerations.}
Each accumulation step is an $N \times N$ matrix product costing $2N^3$ FLOPs, whereas one self-attention layer costs at least $4N^2d$ FLOPs for $QK^\top$ and $AV$ alone, plus $8Nd^2$ for the projections.
The step is therefore cheaper whenever $N < 2d$, which holds for every model in this paper ($N \le 577$ per image or crop; the tightest case is DeiT-Tiny with $N = 197$ and $d = 192$).
Appendix~\ref{app:scaling} measures how the two compare for single sequences up to $N = 4096$.

\subsection{End-to-End Inference Latency}\label{appendix:latency}

The overhead analysis above characterizes the cost of the accumulation and clustering stages in isolation. Here we evaluate the \textit{end-to-end} wall-clock impact, demonstrating that ASAP's early compression yields a net latency reduction compared to methods that defer token reduction to later layers.

As reported in Table~\ref{tab:latency_comparison} (Section~\ref{sec:latency}), ASAP achieves the fastest average inference time (0.064s) and prefill time (0.035s) on the POPE benchmark under a 64-token budget. The key structural reason is that ASAP executes token reduction at $t^*$ (typically layers 12--14 for CLIP ViT-Large with $L=24$), whereas late-pruning methods such as VisionZip~\cite{yang2025visionzip} defer compression until the penultimate layer. This means ASAP eliminates background tokens before they incur the quadratic attention cost at layers $t^*+1, \ldots, L$, and further reduces the sequence length processed by the LLM decoder. The resulting downstream savings far exceed the nominal accumulation overhead quantified in Section~\ref{appendix:overhead_breakdown}, yielding a 10\% prefill latency reduction compared to VisionZip.

PruneSID~\cite{fang2026prune} applies SVD-based PCA grouping followed by iterative NMS with pairwise comparisons, resulting in slightly higher prefill latency (41\,ms) than VisionZip (39\,ms). In contrast, ASAP computes $N$ scalar distances from a single reference point followed by $O(N \log N)$ sorting, achieving the lowest prefill time (35\,ms) among all methods. More importantly, at 64 tokens ASAP reaches 85.7\% POPE accuracy, close to the full 576-token baseline (85.9\%), compared with 84.1\% for PruneSID and 77.0\% for VisionZip, so the latency advantage does not come at the cost of accuracy.

\paragraph{Scalability to LLaVA-NeXT ($N{=}2880$).}

LLaVA-NeXT encodes each image into up to 2880 tokens via the AnyRes strategy (5 crops $\times$ 576 tokens per crop), where each crop is processed independently through the CLIP ViT encoder. While the per-crop accumulation cost remains at $N=576$ scale, the total visual token sequence reaching the LLM decoder grows to 2880, making decoder-side compression critical.

\begin{wraptable}{r}{0.5\textwidth}
    \centering
    \vspace{-10pt} % 상단 여백 조절
    \caption{End-to-End Latency Comparison of LLaVA-NeXT.}
        \label{tab:latency_comparison_next}
        \resizebox{\linewidth}{!}{
        \begin{tabular}{lcccc}
            \toprule
            Method & Tokens & Inference (ms) & Prefill (ms) & Acc (\%) \\
            \midrule
            LLaVA-NeXT (Full)         & 2880 & 268 & 234 & 86.4 \\
            VisionZip            & 160  & 115 & 89 & 74.8 \\
            PruneSID             & 160  & 123 & 105 & 76.9 \\
            \textbf{ASAP (Ours)} & \textbf{160} & \textbf{98} & \textbf{78} & \textbf{81.6} \\
            \bottomrule
        \end{tabular}
        }
    \vspace{-10pt}
\end{wraptable}

As reported in Table~\ref{tab:latency_comparison_next}, ASAP achieves 98\,ms average inference and 78\,ms prefill under a 160-token budget on LLaVA-NeXT, reducing prefill latency by 67\% compared to the full model (234\,ms) and by 12\% compared to VisionZip (89\,ms). The gap over PruneSID is more pronounced: PruneSID's SVD-based grouping over 2880-dimensional token matrices incurs 105\,ms prefill---35\% slower than ASAP. This widening gap compared to the LLaVA-1.5 setting (41\,ms vs.\ 35\,ms, a 17\% difference) confirms that ASAP's $O(N^2)$ distance computation scales more favorably than PruneSID's $O(N^2 d)$ pairwise comparisons as $N$ increases.

Crucially, the accuracy advantage is even more decisive at this scale: ASAP achieves 81.6\% POPE accuracy versus 76.9\% for PruneSID and 74.8\% for VisionZip, representing a 4.7\%p and 6.8\%p margin respectively. Combined with the latency reduction, this demonstrates that the accumulation overhead of the cumulative transition matrix remains negligible relative to the downstream savings from early token compression, even at $N{=}2880$.

\subsection{Modular Computational Cost Analysis}\label{appendix:cca}

\begin{table}[h]
    \centering
    \caption{Step-by-step computational cost breakdown of ASAP on LLaVA-NeXT ($N=2880$). Memory is reported in GB.}
    \label{tab:llava_next_breakdown}
    \begin{tabular}{lcccc}
    \toprule
    Method & Tokens & FLOPs (T) & GPU Memory & CUDA Time (ms) \\
    \midrule
    LLaVA-NeXT & 2880 & 43.6 & 17.2 & 311 \\
    + Lazy Random Walk (All layers) & 2880 & 44.0 & 17.3 & 340 \\
    + $\tau$ detect stop (Early stopping) & 2880 & 43.7 & 17.2 & 319 \\
    + ASAP $C_1$ Pooling & avg ($\sim$1050) & 20.1 & 15.4 & 183 \\
    + ASAP Stage 2 & 160 & 3.8 & 14.0 & 42 \\
    \bottomrule
    \end{tabular}
\end{table}

For high-resolution models where the number of visual tokens increases significantly (e.g., $N=2880$ in LLaVA-NeXT), there may be concerns that the $O(N^3)$ matrix multiplication required for the Lazy Random Walk accumulation could become a computational bottleneck. To verify this, we meticulously measured the step-by-step computational cost of ASAP on LLaVA-NeXT (Table \ref{tab:llava_next_breakdown}).

Our measurements reveal that accumulating the transition matrix across all layers introduces a latency overhead of approximately 29ms (CUDA Time) compared to the original baseline. However, by applying our core $\tau$-based early stopping mechanism, this overhead is strictly bounded to a mere 8ms (with FLOPs marginally increasing from 43.6T to 43.7T). 

More importantly, this minimal initial investment translates into overwhelming downstream computational savings. Immediately after the $C_1$ Pooling stage---which compresses the geometric background at layer $t^*$---the number of tokens drops to an average of $\sim$1050, significantly reducing the total latency to 183ms. When applying the extreme compression of Stage 2 (160 tokens), the total computational cost plummets to 3.8T FLOPs and 42ms in latency, achieving a $>7\times$ reduction in computational cost compared to the full baseline. This demonstrates that constructing the cumulative transition matrix, when paired with early stopping, is a highly practical and scalable architecture even in high-resolution environments.

\subsection{Scaling to Long Single Sequences}\label{app:scaling}

Our LLaVA-NeXT experiments accumulate per crop ($N{=}576$), so they do not exercise single sequences with $N > 1000$.
We therefore benchmark the two kernels in isolation: one accumulation step ($P \leftarrow P\tilde{A}$, $N{\times}N$) and one eager self-attention layer (LayerNorm, QKV projection, $QK^\top$, softmax, $AV$, output projection; $d{=}1024$, 16 heads).
The attention layer excludes the MLP block, which understates the denominator and is therefore conservative.
Eager attention is used because ASAP needs the explicit attention map at layers $\le t^*$.
Timings are medians over 100 runs with CUDA events on an NVIDIA A100 80GB (PyTorch 2.1.2).
``Chained'' is the measured time of $t^*{-}1{=}11$ consecutive products ($t^*{=}12$), which can be below 11$\times$ the single step due to kernel pipelining.
Because the precision policy changes absolute times, we report all three settings.

\begin{table}[h]
\centering
\small
\setlength{\tabcolsep}{4.5pt}
\caption{Accumulation versus one attention layer for a single sequence of length $N$ ($d{=}1024$). Times in ms; ratio = accumulation step / attention layer. Achieved TFLOPS use $2N^3$ for accumulation and $8Nd^2 + 4N^2d$ for attention.}
\label{tab:scaling}
\begin{tabular}{llcccccc}
\toprule
Precision & $N$ & Accum.\ step & Attn.\ layer & Ratio & Chained ($t^*{=}12$) & \multicolumn{2}{c}{TFLOPS (accum.\ / attn.)} \\
\midrule
\multirow{4}{*}{TF32}
 & 576  & 0.024 & 0.242  & 0.10 & 0.15  & 16.2  & 25.6 \\
 & 1024 & 0.044 & 0.469  & 0.09 & 0.37  & 48.8  & 27.5 \\
 & 2048 & 0.183 & 1.390  & 0.13 & 2.11  & 93.7  & 24.7 \\
 & 4096 & 1.079 & 4.864  & 0.22 & 13.13 & 127.4 & 21.2 \\
\midrule
\multirow{4}{*}{fp16}
 & 576  & 0.024 & 0.251  & 0.09 & 0.10  & 16.2  & 24.7 \\
 & 1024 & 0.038 & 0.337  & 0.11 & 0.18  & 56.7  & 38.2 \\
 & 2048 & 0.087 & 0.962  & 0.09 & 0.85  & 197.4 & 35.7 \\
 & 4096 & 0.546 & 2.872  & 0.19 & 6.44  & 251.6 & 35.9 \\
\midrule
\multirow{4}{*}{\shortstack[l]{fp32\\(no TF32)}}
 & 576  & 0.042 & 0.600  & 0.07 & 0.35  & 9.1   & 10.3 \\
 & 1024 & 0.141 & 1.328  & 0.11 & 1.44  & 15.2  & 9.7 \\
 & 2048 & 0.985 & 3.513  & 0.28 & 10.73 & 17.4  & 9.8 \\
 & 4096 & 7.316 & 10.815 & 0.68 & 80.36 & 18.8  & 9.5 \\
\bottomrule
\end{tabular}
\end{table}

One accumulation step stays below one attention layer at every $N$ and precision.
At $N{=}4096$, which corresponds to a SAM-scale encoder ($1024^2$ input, $16{\times}16$ patches), the step costs 22\% of an attention layer under TF32, 19\% in fp16, and 68\% in strict fp32.
The full accumulation to $t^*{=}12$ amounts to 11\%, 9\%, and 31\% of the 24 attention layers, respectively (116.7, 68.9, and 259.6\,ms).
This holds even though the accumulation already exceeds the attention layer in FLOPs at $N{=}4096$ (137 vs.\ 103 GFLOPs, $1.33\times$).
The achieved-TFLOPS columns explain the gap: the square matrix product runs close to peak (96\% of the 19.5 TFLOPS fp32 peak; 81\% of the 312 TFLOPS fp16 peak), while the attention layer reaches 49\% and 12\%, respectively, because softmax, LayerNorm, and the $N{\times}N$ map are memory-bound.
These are kernel-level measurements; the end-to-end anchor for a real model remains Table~\ref{tab:llava_next_breakdown} ($N{=}2880$ across crops).

\section{Modular Composition Analysis}\label{appendix:hybrid}

\begin{table}[H]
    \centering
    \caption{Hybrid ablation study on LLaVA. Starting from adaptive $C_1$-only pooling ($\sim$230 tokens), we compare different Stage 2 compression strategies to reach a fixed 64-token budget.}
    \label{tab:vlm_ablation}
    \resizebox{\linewidth}{!}{
        \begin{tabular}{lccccccccccc}
            \toprule
            \multirow{2}{*}{Method} & \multirow{2}{*}{\begin{tabular}[c]{@{}c@{}}Remain \\ Tokens\end{tabular}} & \multicolumn{7}{c}{Benchmarks ($\uparrow$)} & \multirow{2}{*}{Avg.} & \multirow{2}{*}{Rel (\%)} \\
            \cmidrule(lr){3-9}
            & & GQA & MME & POPE & SQA-Img & VizWiz & VQAv2 & MMB-En & \\
            \midrule
            LLaVA (Full) & 576 & 61.9 & 1862 & 85.9 & 69.5 & 54.3 & 78.5 & 64.7 & 72.6 & 100 \\
            \midrule
            Only $C_1$ Pooling & adaptive (avg. 230) & 61.7 & 1832 & 86.9 & 69.1 & 56.0 & 76.6 & 64.0 & 72.3 & 99.5 \\
            \midrule
            + VisPruner & 64 & 58.9 & 1712 & 84.8 & 67.9 & 54.9 & 72.8 & 59.7 & 69.3 & 95.4 \\
            + VisPruner + Transition Accumulation & 64 & 59.2 & 1716 & 85.1 & 68.1 & 54.6 & 72.8 & 60.3 & 69.4 & 95.6 \\
            \midrule
            + VisPruner (Diffusion Dist) & 64 & 59.5 & 1771 & 85.0 & 68.1 & 54.1 & \textbf{73.2} & 61.3 & 69.9 & 96.3 \\
            \textbf{+ ASAP + VisPruner (ASAP Hybrid)} & 64 & \textbf{59.6} & \textbf{1781} & \textbf{85.7} & \textbf{68.3} & \textbf{56.0} & 73.0 & \textbf{62.0} & \textbf{70.2} & \textbf{96.6} \\
            \midrule
            \textbf{ASAP (Original)} & 64 & 58.1 & 1697 & 84.7 & 68.6 & 56.0 & 72.5 & 61.5 & 69.5 & 95.7 \\
            \bottomrule
        \end{tabular}
    }
\end{table}

ASAP achieves strong performance as a standalone token reduction framework, but its modular structure also allows it to function naturally as a preprocessing stage for existing methods. In this section, we isolate each component of this composition to analyze how ASAP enhances the performance of downstream reduction methods. Specifically, we use VisPruner~\cite{zhang2025beyond} as the plug-in method for Stage 2 compression, as it provides a representative bipartite-matching-based reduction strategy.

\textbf{Standalone effectiveness of background removal.} As shown in Table~\ref{tab:vlm_ablation}, $C_1$-only pooling—which compresses the background cluster into a single representative token while retaining all foreground clusters as-is—adaptively preserves an average of approximately 230 tokens while achieving 99.5\% relative performance compared to full LLaVA. Notably, the POPE score of 86.9\% exceeds the full baseline (85.9\%), indicating that geometric background removal alone is sufficient to reduce visual noise for the LLM decoder. This result confirms that ASAP's core contribution—sink-anchored foreground–background separation—is effective independently of any downstream method.

\textbf{ASAP as a standalone 64-token method.} To reach the 64-token budget without any external plug-in, ASAP applies diffusion-distance-based stride sampling to the surviving foreground clusters after $C_1$ pooling. This achieves 69.5 avg (95.7\% relative), already matching or exceeding all compared baselines in Table~\ref{tab:vlm_reduction_results}, confirming that the core pipeline alone is competitive at extreme compression.

\textbf{Effect of metric choice in the plug-in stage.} When combining $C_1$ pooling with VisPruner for further compression to 64 tokens, the choice of similarity metric plays a decisive role. Using VisPruner with its original cosine similarity metric achieves 95.4\% relative performance, whereas replacing cosine similarity with diffusion distance as the matching metric improves this to 96.3\%---a gap of 0.9\%p. This difference arises because diffusion distance measures information flow independence on the global manifold, rather than the angular difference between local feature vectors that cosine similarity captures. In other words, the geometric information encoded in the cumulative transition matrix $P^{(t^*)}$ serves as a consistently more powerful metric not only for background removal but also for redundancy identification in the plug-in stage.

\textbf{Generality of cumulative transition accumulation.} Interestingly, even within VisPruner's original cosine-similarity-based matching, applying the Markov-chain cumulative transition matrix yields a marginal improvement over single-layer attention (95.4\% $\to$ 95.6\%). While the gain itself is small, it suggests that reflecting global context is beneficial regardless of the metric used. However, cosine similarity cannot fully exploit the rich geometric structure of the accumulated matrix, whereas diffusion distance is inherently designed to do so---explaining the synergistic effect observed in the full ASAP Hybrid variant (96.6\%).

\textbf{Summary.} The analysis above demonstrates three aspects of ASAP's flexibility. First, $C_1$-only pooling is self-sufficient for moderate compression (128 tokens and above), achieving 99.5\% relative performance by background removal alone. Second, even at extreme compression (64 tokens), ASAP's standalone pipeline achieves 95.7\% relative performance, competitive with all baselines without requiring any external method. Third, when combined with existing methods such as VisPruner as a preprocessing stage, sharing diffusion distance as the metric across both stages yields the best overall result (96.6\%), confirming that ASAP's geometric framework consistently enhances downstream reduction methods.

\section{Anchor Selection Ablation on LLaVA-NeXT}\label{appendix:anchor_selection_llavanext}

\begin{table}[t]
\centering
\caption{Anchor selection ablation on LLaVA-NeXT-7B at 160 visual 
tokens. All variants share identical cumulative transition matrix 
accumulation, diffusion distance computation ($\ell_2$ norm between 
$P^{(t^*)}$ rows), Radial Diffusion Clustering, and Transition Weight 
Pooling—only the anchor selection strategy differs. PCA center is 
computed via \texttt{torch.pca\_lowrank} (q=$K$) on $P^{(t^*)}$, with 
the anchor set to the token nearest to the projected mean. Best 
results in \textbf{bold}. }
\label{tab:vlm_anchor_ablation}
\small
\begin{tabular}{lccccc}
\toprule
Anchor & GQA $\uparrow$ & POPE $\uparrow$ & MME $\uparrow$ & SQA-Img $\uparrow$ & Avg. Rel (\%) $\uparrow$ \\
\midrule
Random          & 53.3 & 79.6 & 1481 & 66.9 & 87.5 \\
PCA Center      & 53.1 & 75.2 & 1564 & 67.3 & 87.5 \\
\textbf{Attention Sink (Ours)} & \textbf{57.0} & \textbf{81.6} & \textbf{1730} & \textbf{67.8} & \textbf{93.6} \\
\bottomrule
\end{tabular}
\end{table}

To verify that the attention sink uniquely serves as an effective geometric anchor, we compare against two alternative anchor selection 
strategies under identical pipelines (cumulative transition matrix, diffusion distance, Radial Clustering, Transition Weight Pooling): 
(i) a randomly chosen token, and (ii) the token nearest to the PCA center in the projected transition row space. As reported in 
Table~\ref{tab:vlm_anchor_ablation}, the sink consistently outperforms both alternatives across all four benchmarks, with margins of 
+3.7--3.9 on GQA, +2.0--6.4 on POPE, +166--249 on MME, and +0.5--0.9 on SQA-Img.

A particularly notable finding is that PCA center underperforms even random selection on POPE (75.2 vs. 79.6). We attribute this to a 
systematic bias of statistical centrality toward salient foreground regions: foreground tokens contribute disproportionately to feature 
variance, pulling the PCA center toward semantically meaningful objects. When such tokens serve as the anchor, foreground tokens 
yield near-zero diffusion distances and are erroneously compressed as background---a failure mode analogous to the random-anchor 
miscompression illustrated in Figure~\ref{appendix:negative_ra}, but occurring \emph{systematically} rather than stochastically. 
This explains the particularly large gap on POPE, which directly penalizes object-level information loss through hallucination metrics.

In contrast, the sink's position is determined by information flow convergence rather than feature distribution, ensuring placement in 
semantically neutral regions regardless of foreground prominence. This robustness validates our central thesis that the attention 
sink---a \emph{dynamical} attractor---is fundamentally distinct from \emph{statistical} centroids, despite both being interpretable 
as ``central'' tokens.

\section{Sink Emergence Dynamics Across Architectures}\label{appendix:sink_emergence}

In this section, we provide both qualitative and quantitative analysis of how the attention sink forms across layers in different architectures, validating the robustness of the threshold-based detection mechanism.

\begin{figure}[H]
    \centering
    \begin{subfigure}[b]{0.9\linewidth}
        \centering
        \includegraphics[width=\linewidth]{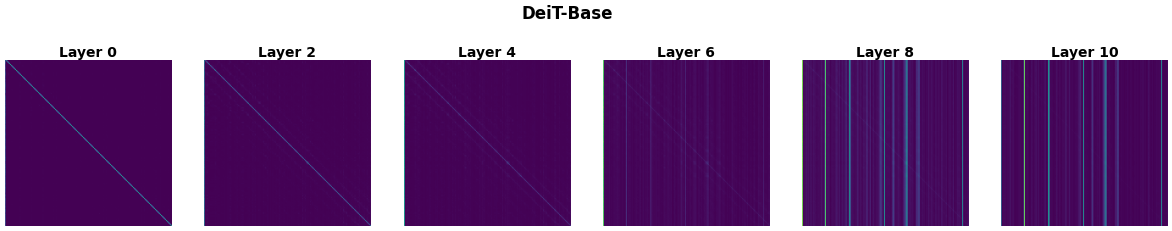}
    \end{subfigure}
    \begin{subfigure}[b]{0.9\linewidth}
        \centering
        \includegraphics[width=\linewidth]{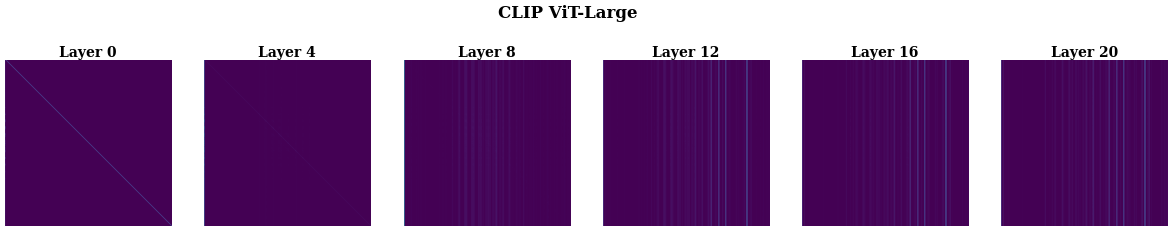}
    \end{subfigure} 
    \begin{subfigure}[b]{0.9\linewidth}
        \centering
        \includegraphics[width=\linewidth]{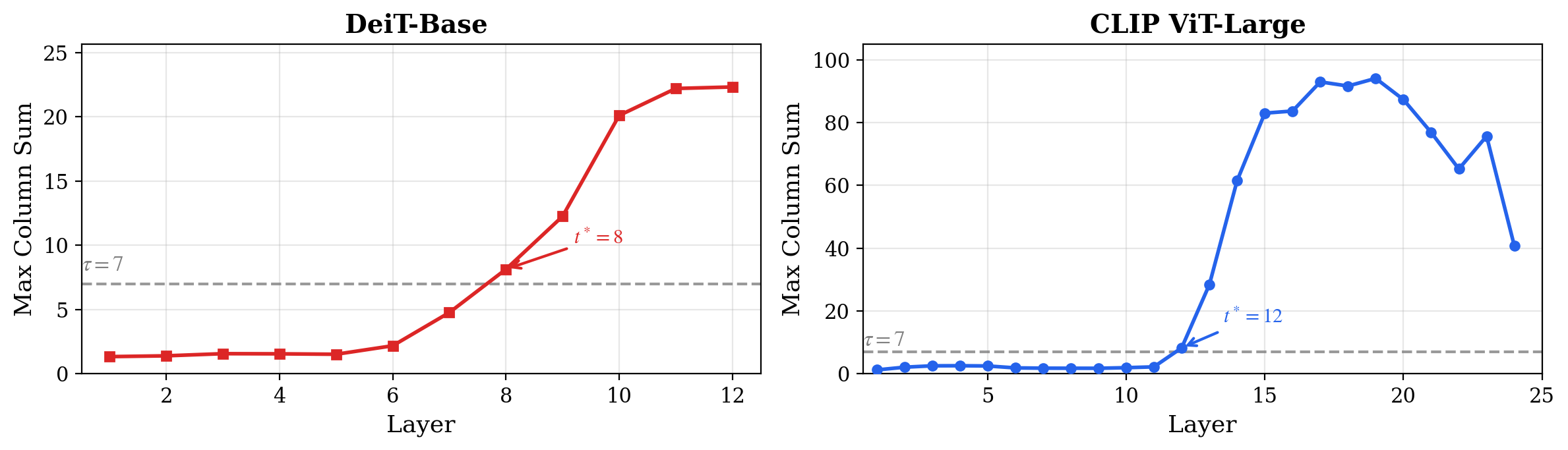} 
    \end{subfigure}
    \caption{Sink emergence dynamics for DeiT-Base (L=12, N=197) and CLIP ViT-Large (L=24, N=577).(Top two rows) Cumulative transition matrix $P^{(t)}$ visualized at selected layers, labeled with 0-based layer indices (Layer~$\ell$ shows $P^{(\ell+1)}$). In early layers, the diagonal structure dominates, indicating that each token primarily retains its own information. As layers deepen, probability mass progressively concentrates into a sparse set of columns (bright vertical stripes), corresponding to the formation of attention sinks as mass-accumulating attractors. \textbf{(Bottom)} Max column sum of $P^{(t)}$ across layers. Both architectures exhibit a sharp phase transition from a flat baseline ($\sim$2) to rapid accumulation, despite substantial differences in depth and token count. The threshold $\tau \in \{7, 8\}$ (dashed line) consistently captures the onset of this transition, detecting sink emergence at $t^*=8$ for DeiT-Base and $t^*=12$ for CLIP ViT-Large.}
    \label{fig:sink_emergence}
\end{figure}

\textbf{Visualization of the cumulative transition matrix.} Figure~\ref{fig:sink_emergence} (top two rows) visualizes the cumulative transition matrix $P^{(t)}$ at selected layers for DeiT-Base and CLIP ViT-Large. In both models, early layers are dominated by the diagonal component, reflecting each token's tendency to preserve its own identity through the residual connection. As the Lazy Random Walk progresses, the diagonal gradually fades and probability mass concentrates into specific columns, forming bright vertical stripes. These stripes correspond to the attention sink: columns where all tokens route a disproportionate share of their information, visually confirming the mass-accumulating attractor interpretation of Appendix~\ref{appendix:proof1}.

\textbf{Phase transition in max column sum.} Figure~\ref{fig:sink_emergence} (bottom) quantitatively tracks the max column sum across layers. Both models maintain a flat baseline of approximately 2 in early layers, followed by a sharp phase transition at a model-specific layer. For DeiT-Base, this transition occurs around layer 8, while for CLIP ViT-Large it occurs around layer 12. Crucially, a threshold in the range $\tau \in \{7, 8\}$ captures the onset of this transition in both cases---detecting the point where the sink structure is sufficiently formed to provide discriminative geometric contrast, yet the manifold has not fully collapsed.

\textbf{Interpretation of $\tau$'s robustness across $N$ and $L$.} The max column sum reflects the cumulative attention mass directed toward the sink by all tokens. Since each token's initial contribution to any given column is approximately $1/N$ under near-uniform early-layer attention, and the column sum starts at approximately 1 regardless of $N$, the absolute magnitude of the max column sum does not scale proportionally with the number of tokens. This means $\tau$ fundamentally measures the \emph{degree of attention concentration} rather than being a function of sequence length. % Empirically, this is confirmed by the fact that the same narrow range $\tau \in \{7, 8\}$ provides an appropriate detection point for both DeiT-Base ($N=197$) and CLIP ViT-Large ($N=577$), suggesting that this threshold transfers across the tested architectures without re-tuning, though validation on a broader range of models is needed.
Empirically, the same narrow range $\tau \in \{7, 8\}$ provides an appropriate detection point for both DeiT-Base ($N=197$) and CLIP ViT-Large ($N=577$), suggesting that $\tau$ depends on attention concentration dynamics rather than scaling directly with $N$. While this means $\tau$ does not need to be re-derived from scratch for each architecture, modest per-architecture tuning within a small window ($\tau \in \{7, 8, 9, 10\}$ is recommended to optimize the accuracy–efficiency trade-off.

\section{Hyperparameter Sensitivity on ViT-AugReg}\label{appendix:augreg_ablation}
 
\begin{figure}[h]
    \centering
    \includegraphics[width=0.55\textwidth]{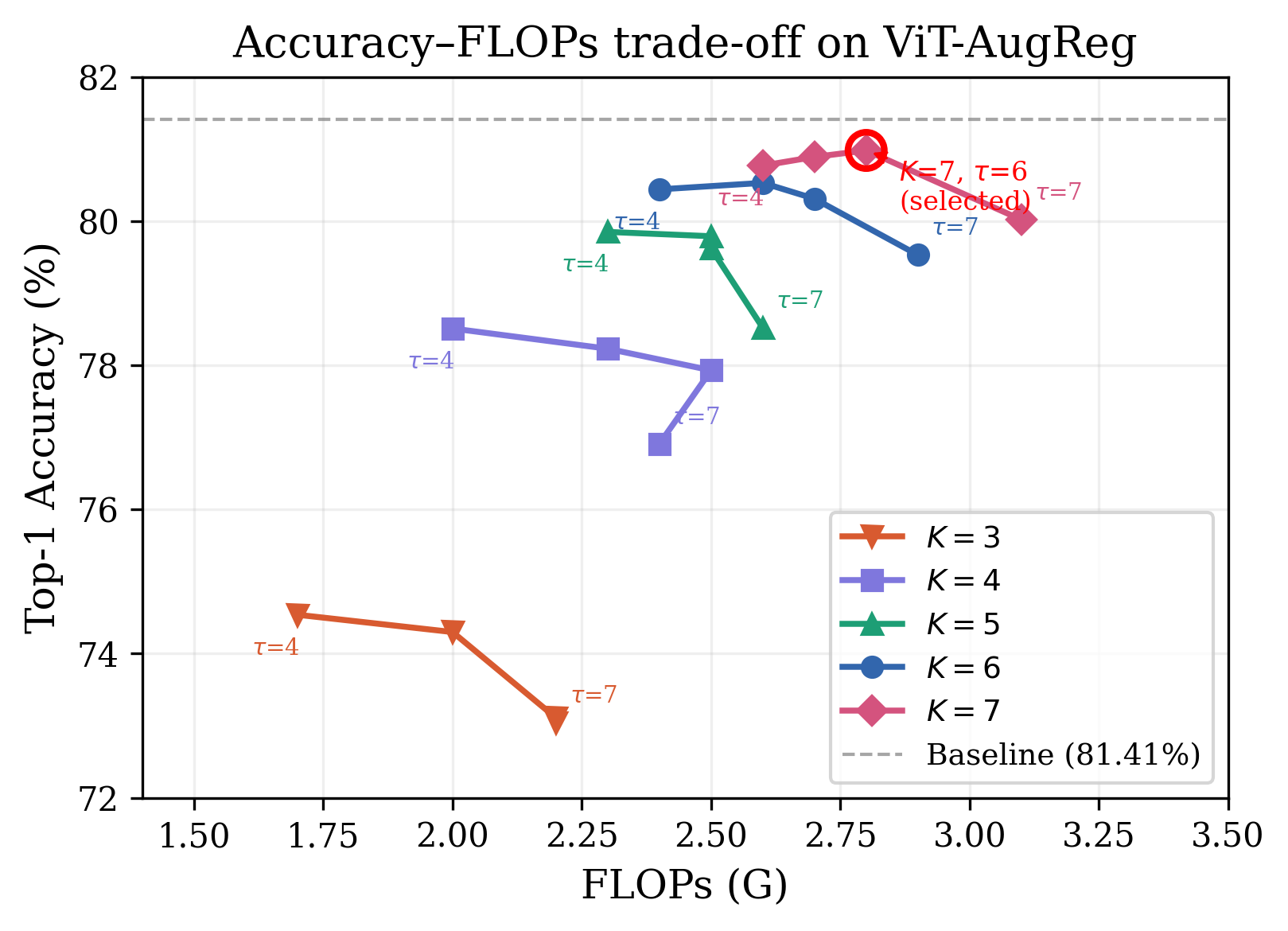}
    \caption{Accuracy--FLOPs trade-off on ViT-AugReg for varying $K$ and $\tau$. The red circle marks the selected operating point ($K{=}7$, $\tau{=}6$).}
    \label{fig:augreg_pareto}
\end{figure}

To verify that the hyperparameter trends observed on DeiT-Base (Section~\ref{subsec:hyperparmas}) generalize across architectures, we conduct the same ablation on ViT-AugReg-B~\cite{steiner2021train} over $K \in \{3,4,5,6,7\}$ and $\tau \in \{4,5,6,7\}$. Figure~\ref{fig:augreg_pareto} summarizes the results.
 
\paragraph{Consistent Lower Bound on $K$.}
As on DeiT-Base, $K{=}3$ is too aggressive: accuracy peaks at 74.54\%, representing a 6.87\%p drop from the baseline. The foreground--background separation boundary remains at $K \geq 6$, where accuracy stabilizes above 79.5\% across all tested $\tau$ values. This confirms that the lower bound on $K$ is not architecture-specific but reflects a general requirement for sufficient bin count to capture the semantic diversity of the diffusion distance distribution.
 
\paragraph{Differences from DeiT-Base.}
Unlike DeiT-Base, where increasing $\tau$ monotonically improves accuracy for all $K$, ViT-AugReg exhibits a non-monotonic pattern for small $K$: at $K{=}3$, accuracy decreases from 74.54\% ($\tau{=}4$) to 73.05\% ($\tau{=}7$). We attribute this to the attention dynamics of AugReg: at deeper layers, the cumulative transition matrix converges more aggressively, causing the diffusion distance distribution to compress. When $K$ is small, this compression merges foreground tokens into the background cluster before sufficient geometric contrast has been preserved. For $K \geq 6$, however, the monotonic trend is recovered, and the best operating point ($K{=}7$, $\tau{=}6$, accuracy 80.98\%) aligns with the configuration reported in Table~\ref{tab:vit_results}.
 
\paragraph{Operating Range.}
The robust operating range on ViT-AugReg is $K \in \{6, 7\}$, $\tau \in \{4, 5, 6\}$, where accuracy varies within 0.67\%p (80.31\%--80.98\%). Combined with the DeiT-Base results, we conclude that $K \in \{5, 6, 7\}$ provides a reliable starting range across architectures, and $\tau$ can be tuned within a small window to control the accuracy--efficiency trade-off.

\section{Lazy Random Walk $\alpha$ Analysis}\label{appendix:alpha}

\begin{table}[h]
    \centering
    \caption{Effect of $\alpha$ on Accuracy and Throughput (DeiT-Base)}
    \label{tab:alpha_ablation}
    \begin{tabular}{ccccc}
    \toprule
    $\alpha$ & Acc (\%) & $\Delta$ Acc & Throughput (img/s) & Throughput $\uparrow$ (\%) \\
    \midrule
    0.1 & 81.44 & $-0.36$ & 1412 & $+16.4$ \\
    0.2 & 81.09 & $-0.71$ & 1387 & $+14.4$ \\
    0.3 & 81.38 & $-0.42$ & 1497 & $+23.4$ \\
    0.4 & 81.61 & $-0.19$ & 1688 & $+39.2$ \\
    0.5 & 81.68 & $-0.12$ & 1799 & $+48.3$ \\
    0.6 & 81.68 & $-0.12$ & 1792 & $+47.8$ \\
    0.7 & 81.57 & $-0.23$ & 1718 & $+41.7$ \\
    0.8 & 80.93 & $-0.87$ & 1863 & $+53.6$ \\
    0.9 & 80.55 & $-1.25$ & 1925 & $+58.7$ \\
    \bottomrule
    \end{tabular}
\end{table}

The parameter $\alpha$ controls the balance between attention-driven information mixing and identity-preserving residual flow in the cumulative transition matrix. When $\alpha$ is too small (e.g., $\alpha \leq 0.3$), the identity component dominates, delaying sink emergence to deeper layers and diluting the geometric contrast in $P^{(t^*)}$. This results in less discriminative diffusion distances and lower accuracy. Conversely, when $\alpha$ is too large (e.g., $\alpha \geq 0.8$), the cumulative matrix converges too rapidly, triggering sink detection before sufficient global context has accumulated — yielding higher throughput but degraded accuracy. The optimal range $\alpha \in \{0.4, 0.5, 0.6\}$ maintains accuracy within 0.19\%p of the baseline while achieving over 39\% throughput improvement. We adopt $\alpha = 0.5$ following the convention of Attention Rollout \citep{abnar2020quantifying}, which also achieves the best accuracy–throughput balance in our experiments.

\section{Qualitative Analysis of Hallucination Suppression}\label{appendix:hallucination}
\begin{figure}[H]
    \centering
    \begin{subfigure}[b]{0.9\linewidth}
        \centering
        \includegraphics[width=\linewidth]{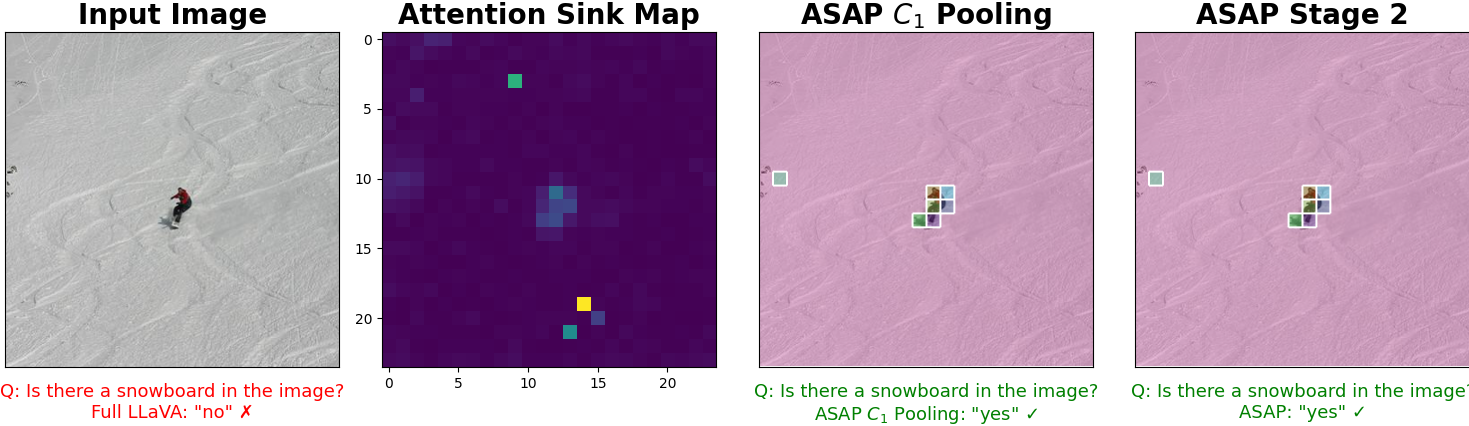}
    \end{subfigure}
    \begin{subfigure}[b]{0.9\linewidth}
        \centering
        \includegraphics[width=\linewidth]{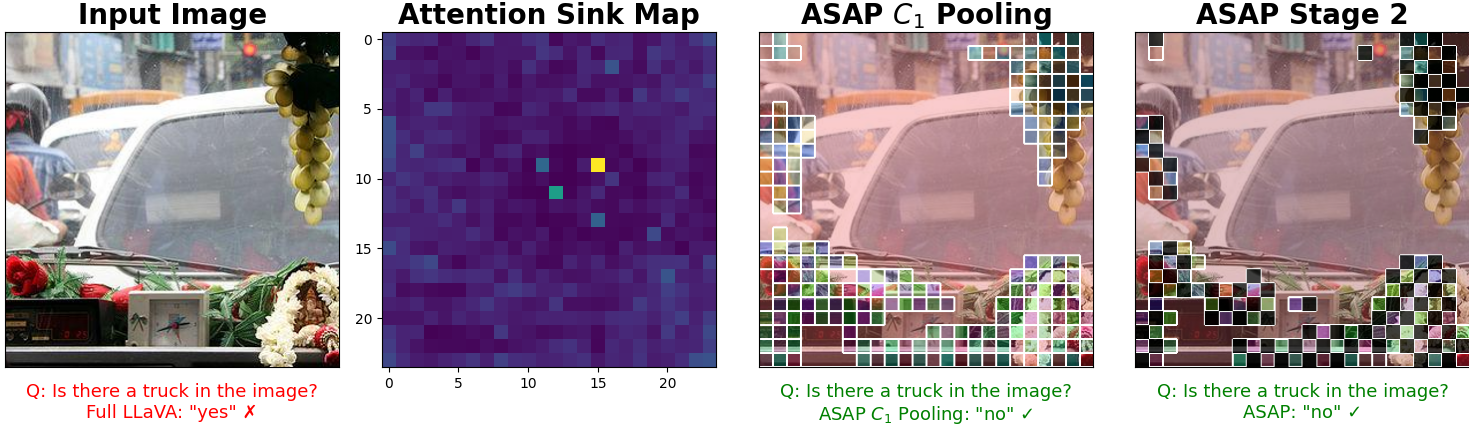}
    \end{subfigure} 
    \begin{subfigure}[b]{0.9\linewidth}
        \centering
        \includegraphics[width=\linewidth]{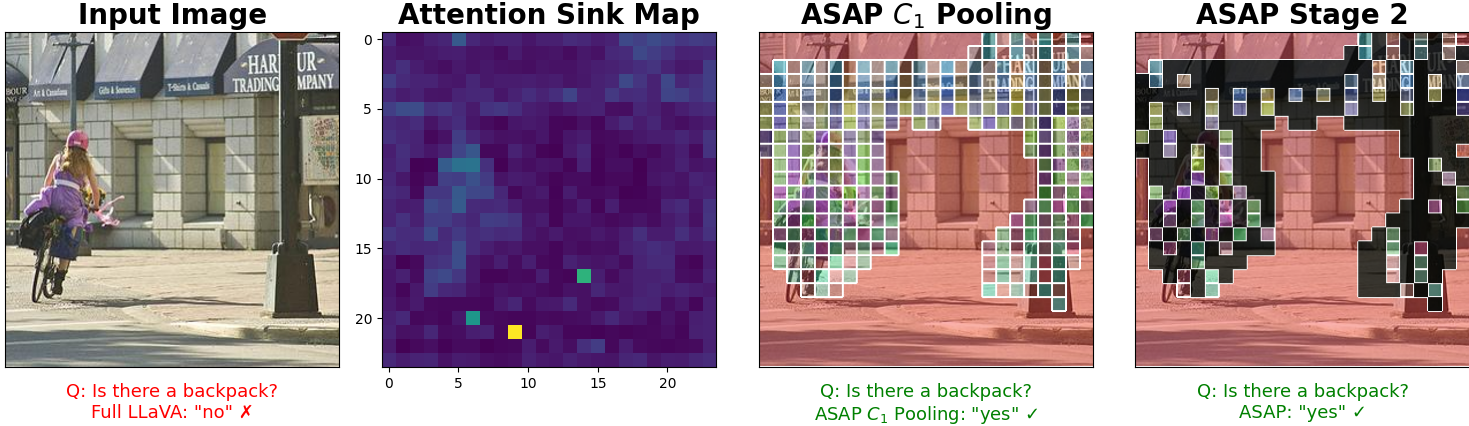} 
    \end{subfigure}
    \begin{subfigure}[b]{0.9\linewidth}
        \centering
        \includegraphics[width=\linewidth]{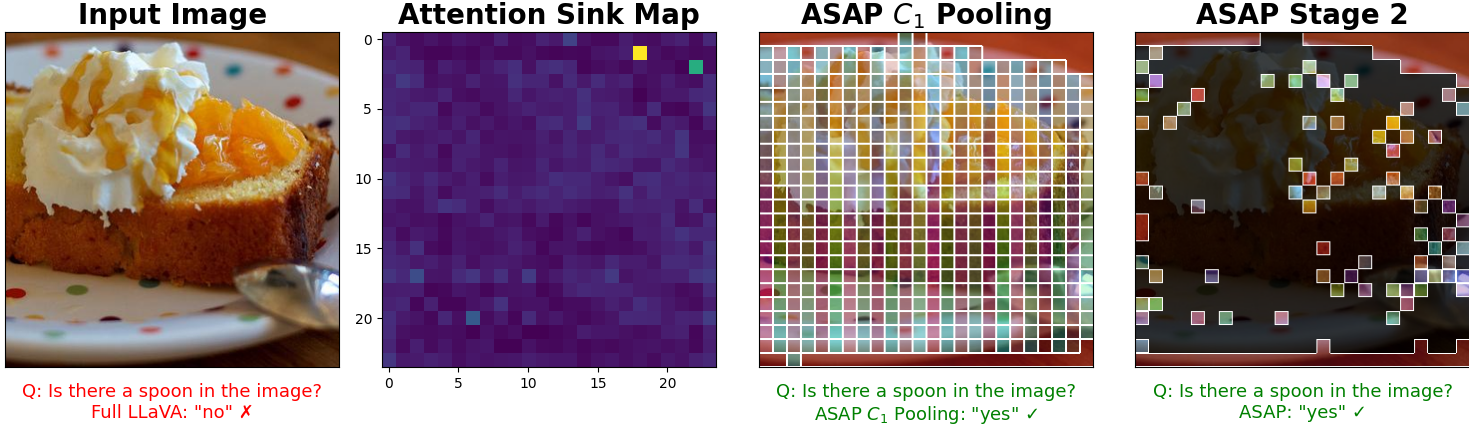} 
    \end{subfigure}
    \caption{Qualitative analysis of hallucination suppression on POPE. Each row shows the input image, attention sink map, tokens retained after $C_1$ Pooling (Stage 1), and the final ASAP output (Stage 2). Red and green 
        borders indicate incorrect and correct predictions, respectively. ASAP corrects four distinct 
        hallucination types: false negatives caused by dominant backgrounds (Row 1), false positives caused 
        by complex visual noise (Row 2), false negatives caused by small occluded foreground objects (Row 3), 
        and false negatives caused by subtle objects on semantically relevant backgrounds (Row 4).}
    \label{fig:pope_qual}
\end{figure}

To qualitatively analyze the POPE score improvement observed in Table~\ref{tab:vlm_ablation}, we present four representative cases in Figure~\ref{fig:pope_qual}. Each case illustrates how a different type of hallucination is corrected through ASAP's background compression.

\textbf{Case 1 (Simple background — FN correction).} In the snowboarder image, the background (snow field) occupies the vast majority of the image. Full LLaVA fails to detect the snowboarder due to the overwhelming background tokens, whereas $C_1$ Pooling alone sufficiently compresses the background and allows the model to focus on the foreground, yielding the correct answer. Notably, in this case, fewer than 64 tokens remain after $C_1$ Pooling, achieving the token budget without requiring Stage 2.

\textbf{Case 2 (Complex background — FP correction).} In a complex rainy street scene, Full LLaVA incorrectly predicts the presence of a truck that does not exist. This false positive arises from visual noise caused by window reflections, rain streaks, and cluttered objects that confuse the LLM decoder. After applying ASAP, background noise tokens are removed and the model correctly predicts the absence of a truck.

\textbf{Case 3 (Small foreground — FN correction).} In the bicycle image, a backpack is a small, partially occluded foreground object. As shown in the Attention Sink Map, the sink forms at semantically neutral pavement and road regions. After $C_1$ Pooling, building wall and pavement tokens are compressed, preserving tokens around the cyclist. Consequently, ASAP correctly detects the backpack that the full model failed to identify.

\textbf{Case 4 (Subtle foreground on semantic background — FN correction).} In the dessert image, a spoon rests on the plate beside the cake. Full LLaVA fails to detect it due to low visual contrast between the spoon and the semantically relevant background (plate surface). After $C_1$ Pooling compresses the dominant plate and cake regions, the spoon tokens are preserved in the retained foreground, yielding a correct detection that persists through Stage 2 compression to 64 tokens.

The four cases cover distinct scenarios — simple background, complex noisy background, small occluded objects, and subtle objects on semantic backgrounds — illustrating how ASAP's geometric background compression can suppress several types of hallucination. Appendix~\ref{app:hallu-quant} complements these cases with a split-wise POPE analysis and with question types where compression loses accuracy.

\subsection{Quantitative Analysis: Where Compression Helps and Hurts}
\label{app:hallu-quant}

\paragraph{POPE by negative-sampling split.}
POPE uses the same positive questions in all three splits and differs only in how the absent objects of the negative questions are sampled (random, popular, adversarial), so differences between splits reflect false positives.
Table~\ref{tab:pope-split} reports F1 per split.

\begin{table}[h]
\centering
\small
\caption{POPE F1 by negative-sampling split.}
\label{tab:pope-split}
\begin{tabular}{lcccc}
\toprule
Model & Random & Popular & Adversarial & Overall \\
\midrule
LLaVA-1.5 (576 tokens)          & 87.28 & 86.17 & 84.21 & 85.87 \\
LLaVA-1.5 + ASAP (128)          & 87.29 & 86.30 & \textbf{85.17} & 86.24 \\
\midrule
LLaVA-NeXT (2880 tokens)        & 87.61 & 86.39 & 85.30 & 86.43 \\
LLaVA-NeXT + ASAP (640)         & 88.28 & 87.25 & \textbf{85.98} & 87.16 \\
\bottomrule
\end{tabular}
\end{table}

ASAP improves over the full model on all three splits for both models.
For LLaVA-1.5, the improvement grows with negative-sampling difficulty ($+0.01$, $+0.13$, $+0.96$) and is concentrated on the adversarial split, where negatives are objects that frequently co-occur with the scene and false positives are most likely.

\paragraph{Losses on VQAv2.}
VQAv2 is where ASAP loses most relative to the full model (Table~\ref{tab:vlm_reduction_results}).
On a fixed subset of 10{,}000 questions, ASAP at 128 tokens scores $-2.22$ points below the full LLaVA-1.5 (95\% bootstrap CI $[-2.7, -1.7]$; McNemar $p < 10^{-14}$).
The loss is not uniform across question types: it is largest for text/OCR ($-6.1$) and color ($-5.7$) questions, and between $-1.3$ and $-1.8$ for existence, counting, and spatial questions.
Pooling the background cluster thus preserves scene-level and object-presence information, consistent with the POPE results above, but can discard small high-frequency details such as text and exact color.
This is a limitation shared by token compression methods generally.

\section{Adaptive Cluster Count: A Preliminary Study}
\label{app:adaptivek}

$K$ is the last manually set hyperparameter.
As a first step toward removing it, we determine the pooled cluster per image instead of using the fixed boundary $\tilde{D}(x_i, x_s) < 1/K$: we apply Otsu's threshold to the distribution of normalized diffusion distances and pool all tokens below it.
On the full ImageNet-val set with DeiT-Base, this reaches 80.49\% top-1, compared with 81.60\% for $K{=}6$ in the same pipeline.

The gap mainly reflects a different operating budget rather than a worse boundary.
The Otsu boundary pools 47\% of the tokens on average, a far more aggressive compression than the $K{=}6$ boundary.
The pooled fraction adapts to the image (10th and 90th percentiles: 34\% and 64\%) and correlates weakly but significantly with the annotation-based background ratio of Appendix~\ref{app:failure} (Spearman $\rho{=}0.17$, $n{=}50{,}000$).

A parameter-free boundary is therefore feasible, but whether it places the boundary better than a fixed $K$ can only be decided at a matched budget.
Comparing the two at equal pooled fractions, and gating Otsu by a bimodality test so that it is applied only to clearly bimodal distance distributions, are left to future work.

\section{Additional Qualitative Results}\label{appendix:additional_plots}

We provide additional qualitative results beyond those shown in the main text. Our method reliably preserves semantically important objects across diverse scene types: both small, easily overlooked foreground objects (e.g., frogs, butterflies) and cases where the object occupies nearly the entire image (e.g., ambulances, car dashboards). This consistency across architectures demonstrates the robustness of our adaptive clustering and pooling mechanism. 
% Furthermore, these additional examples corroborate the observation of \cite{darcet2024vision} that attention sinks consistently form at the least salient regions of the image, regardless of scene composition.
Furthermore, these additional examples are consistent with the observation of~\cite{darcet2024vision} that attention sinks predominantly form at the least salient regions of the image; Appendix~\ref{app:failure} examines the cases where this does not hold.
\subsection{DeiT-Base}\label{appendix:deitbase_plots}

\begin{figure}[H]
    \centering
    % === 첫 번째 줄 (Row 1) ===
    \begin{subfigure}[b]{0.48\linewidth}
        \centering
        \includegraphics[width=\linewidth]{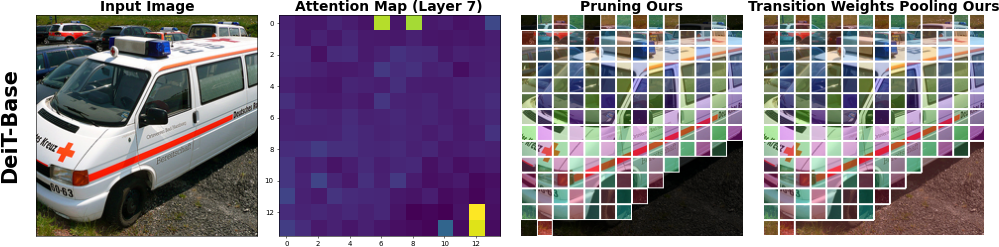}
    \end{subfigure}
    \hfill
    \begin{subfigure}[b]{0.48\linewidth}
        \centering
        \includegraphics[width=\linewidth]{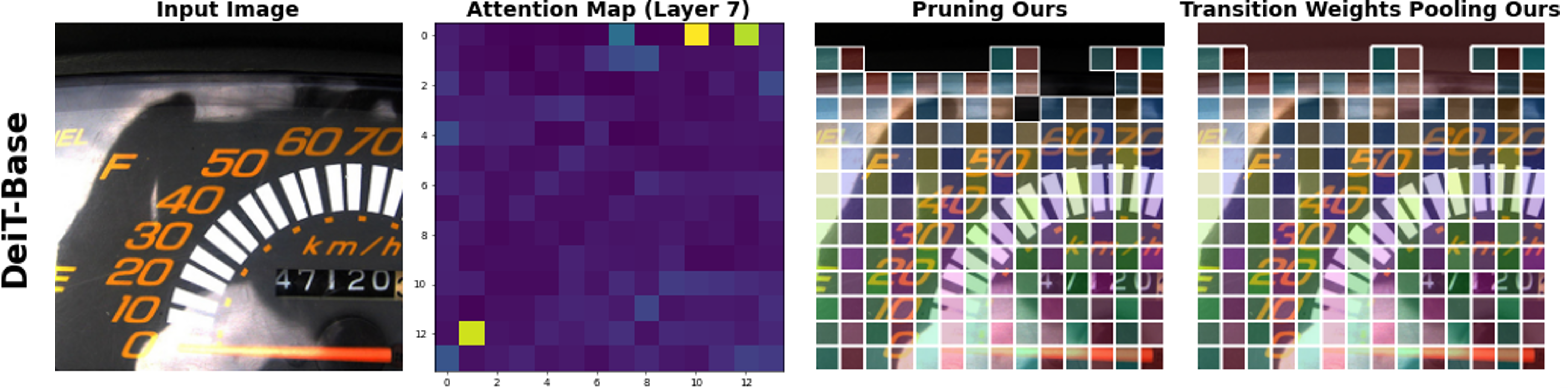}
    \end{subfigure} 
    \begin{subfigure}[b]{0.48\linewidth}
        \centering
        \includegraphics[width=\linewidth]{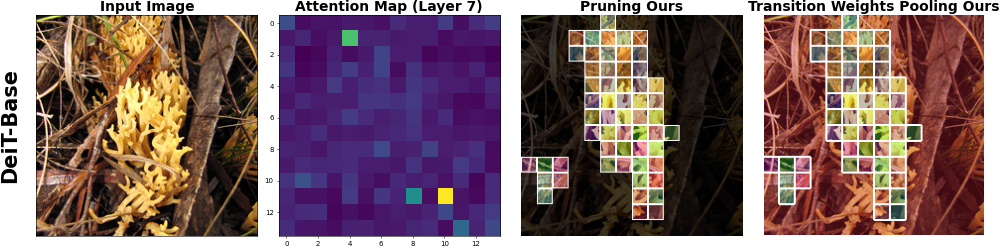} 
    \end{subfigure}
    \hfill
    \begin{subfigure}[b]{0.48\linewidth}
        \centering
        \includegraphics[width=\linewidth]{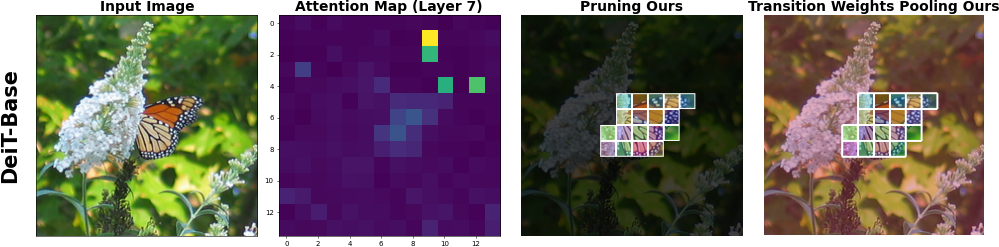}
    \end{subfigure}
    \begin{subfigure}[b]{0.48\linewidth}
        \centering
        \includegraphics[width=\linewidth]{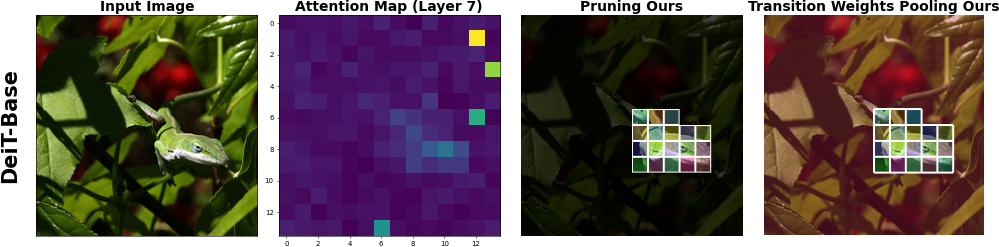} 
    \end{subfigure}
    \hfill
    \begin{subfigure}[b]{0.48\linewidth}
        \centering
        \includegraphics[width=\linewidth]{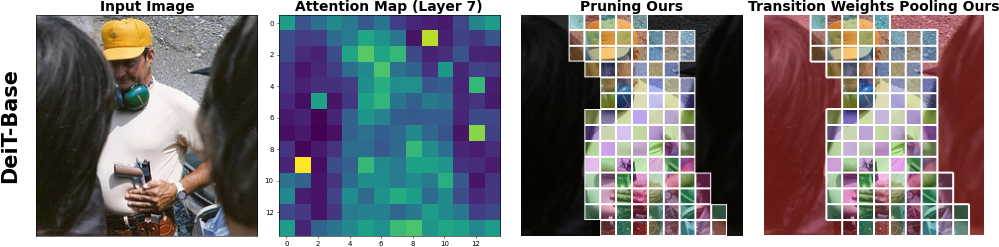}
    \end{subfigure}
    % \begin{subfigure}[b]{0.48\linewidth}
    %     \centering
    %     \includegraphics[width=\linewidth]{appendix/DeiT-base_appendix_figure/appendix_deitb_mogwa.png} 
    % \end{subfigure}
    % \hfill
    % \begin{subfigure}[b]{0.48\linewidth}
    %     \centering
    %     \includegraphics[width=\linewidth]{appendix/DeiT-base_appendix_figure/appendix_deitb_monkey.png}
    % \end{subfigure}
    \begin{subfigure}[b]{0.48\linewidth}
        \centering
        \includegraphics[width=\linewidth]{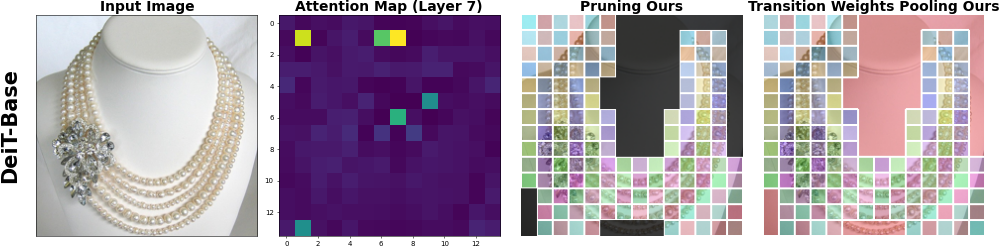} 
    \end{subfigure}
    \hfill
    \begin{subfigure}[b]{0.48\linewidth}
        \centering
        \includegraphics[width=\linewidth]{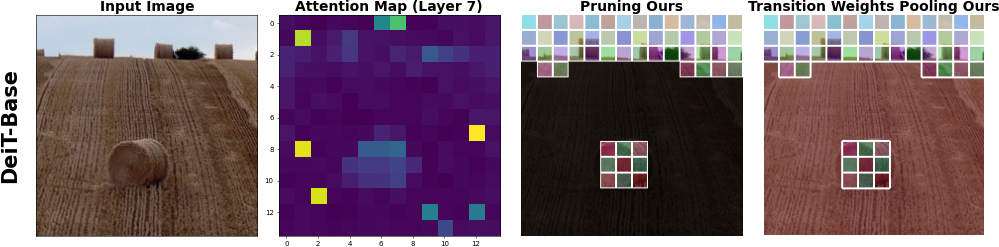}
    \end{subfigure}
    \begin{subfigure}[b]{0.48\linewidth}
        \centering
        \includegraphics[width=\linewidth]{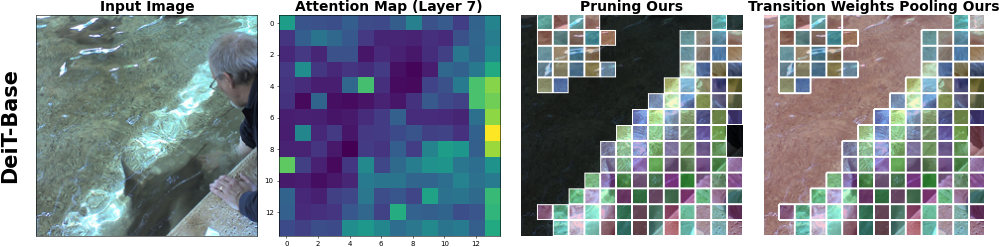} 
    \end{subfigure}
    \hfill
    \begin{subfigure}[b]{0.48\linewidth}
        \centering
        \includegraphics[width=\linewidth]{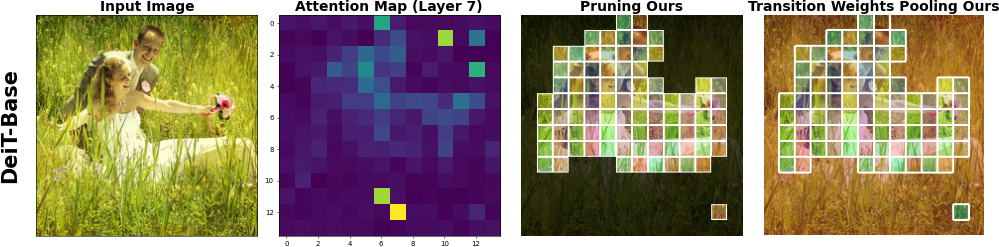}
    \end{subfigure}
    \caption{Qualitative results on DeiT-Base}
\end{figure}

\subsection{ViT-AugReg}\label{appendix:augreg_plots}

\begin{figure}[H]
    \centering
    % === 첫 번째 줄 (Row 1) ===
    \begin{subfigure}[b]{0.48\linewidth}
        \centering
        \includegraphics[width=\linewidth]{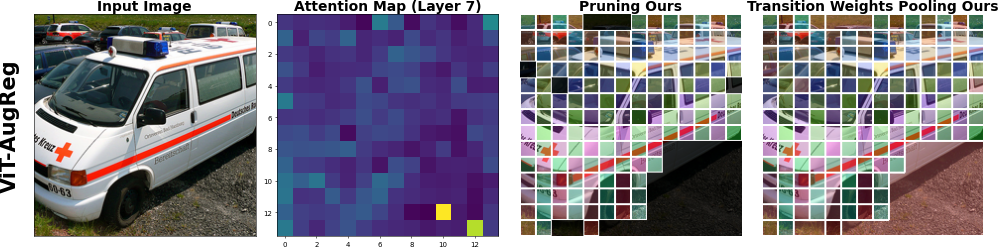}
    \end{subfigure}
    \hfill
    \begin{subfigure}[b]{0.48\linewidth}
        \centering
        \includegraphics[width=\linewidth]{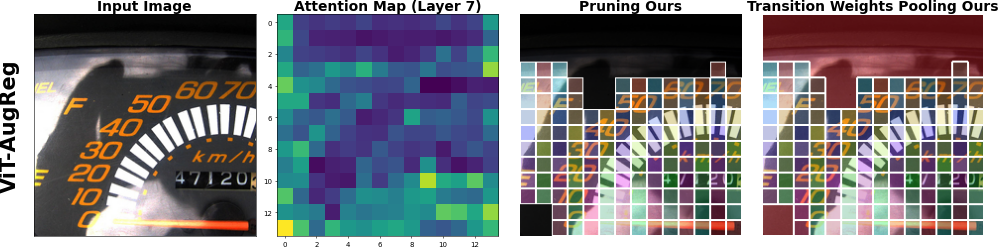}
    \end{subfigure} 
    \begin{subfigure}[b]{0.48\linewidth}
        \centering
        \includegraphics[width=\linewidth]{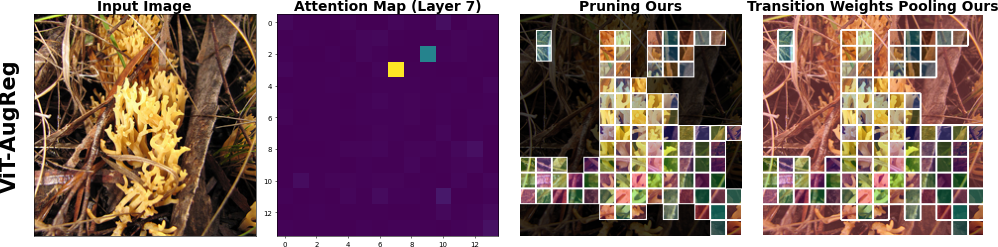} 
    \end{subfigure}
    \hfill
    \begin{subfigure}[b]{0.48\linewidth}
        \centering
        \includegraphics[width=\linewidth]{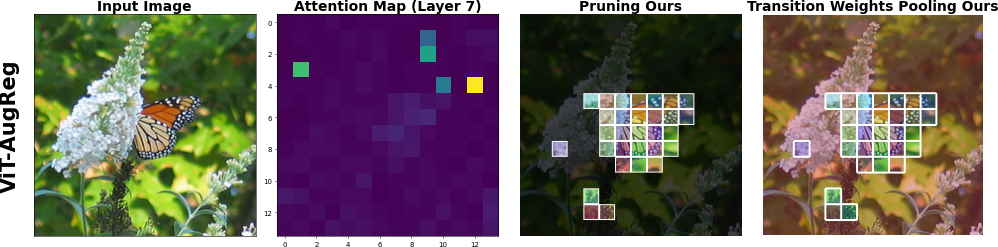}
    \end{subfigure}
    \begin{subfigure}[b]{0.48\linewidth}
        \centering
        \includegraphics[width=\linewidth]{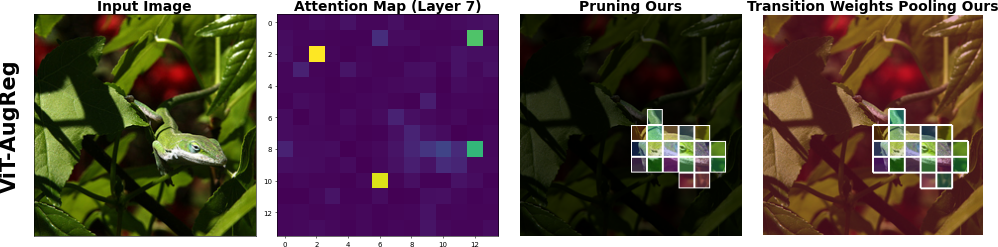} 
    \end{subfigure}
    \hfill
    \begin{subfigure}[b]{0.48\linewidth}
        \centering
        \includegraphics[width=\linewidth]{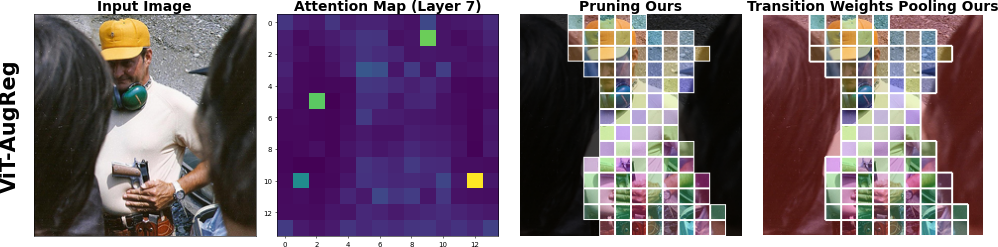}
    \end{subfigure}
    % \begin{subfigure}[b]{0.48\linewidth}
    %     \centering
    %     \includegraphics[width=\linewidth]{appendix/AugReg_appendix_figure/appendix_augreg_mogwa.png} 
    % \end{subfigure}
    % \hfill
    % \begin{subfigure}[b]{0.48\linewidth}
    %     \centering
    %     \includegraphics[width=\linewidth]{appendix/AugReg_appendix_figure/appendix_augreg_monkey.png}
    % \end{subfigure}
    \begin{subfigure}[b]{0.48\linewidth}
        \centering
        \includegraphics[width=\linewidth]{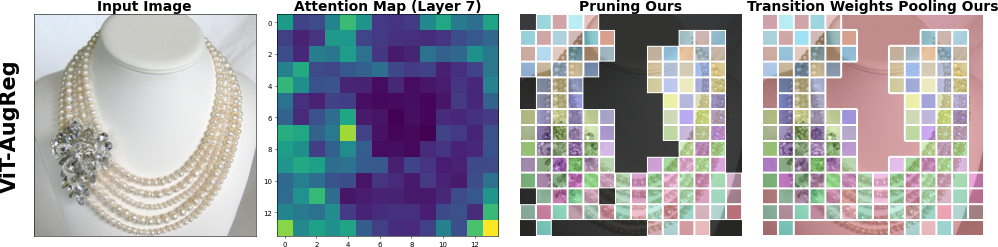} 
    \end{subfigure}
    \hfill
    \begin{subfigure}[b]{0.48\linewidth}
        \centering
        \includegraphics[width=\linewidth]{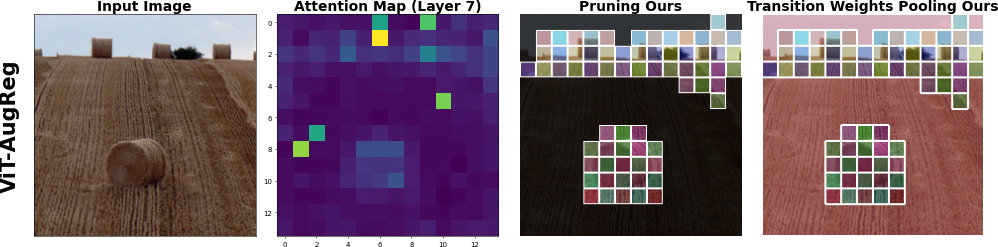}
    \end{subfigure}
    \begin{subfigure}[b]{0.48\linewidth}
        \centering
        \includegraphics[width=\linewidth]{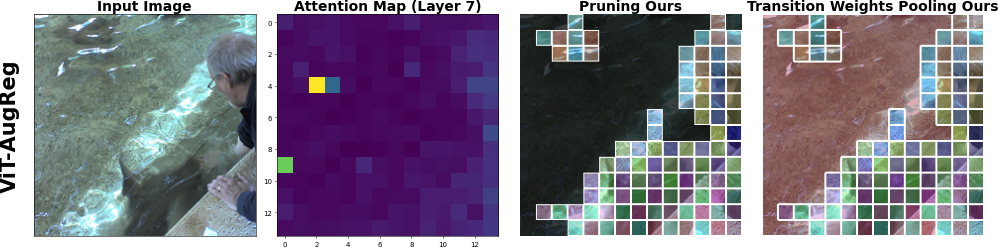} 
    \end{subfigure}
    \hfill
    \begin{subfigure}[b]{0.48\linewidth}
        \centering
        \includegraphics[width=\linewidth]{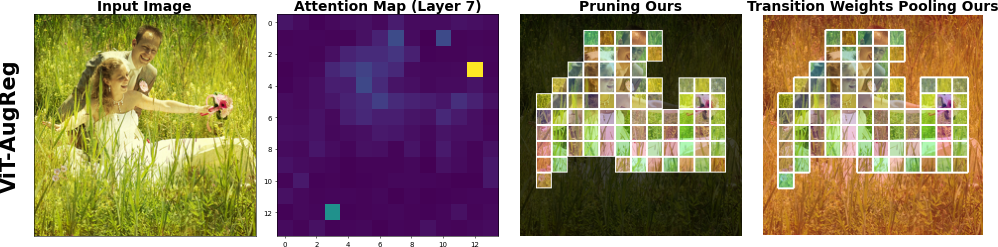}
    \end{subfigure}
    \caption{Qualitative results on ViT-AugReg}
\end{figure}

\subsection{LV-ViT-S}\label{appendix:lvvit_plots}

\begin{figure}[H]
    \centering
    \vspace{-5pt}
    % === 첫 번째 줄 (Row 1) ===
    \begin{subfigure}[b]{0.48\linewidth}
        \centering
        \includegraphics[width=\linewidth]{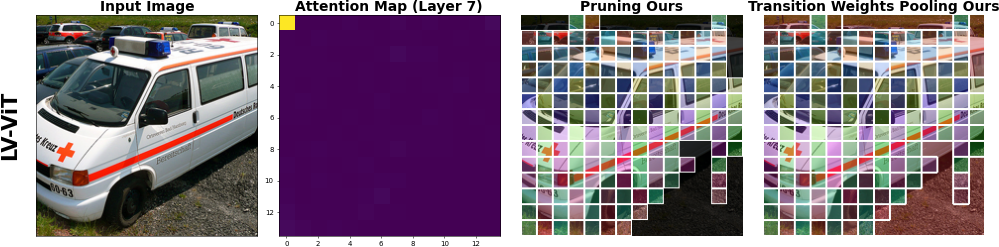}
    \end{subfigure}
    \hfill
    \begin{subfigure}[b]{0.48\linewidth}
        \centering
        \includegraphics[width=\linewidth]{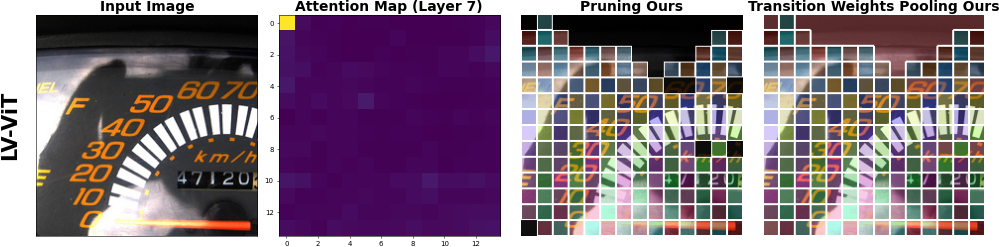}
    \end{subfigure} 
    \begin{subfigure}[b]{0.48\linewidth}
        \centering
        \includegraphics[width=\linewidth]{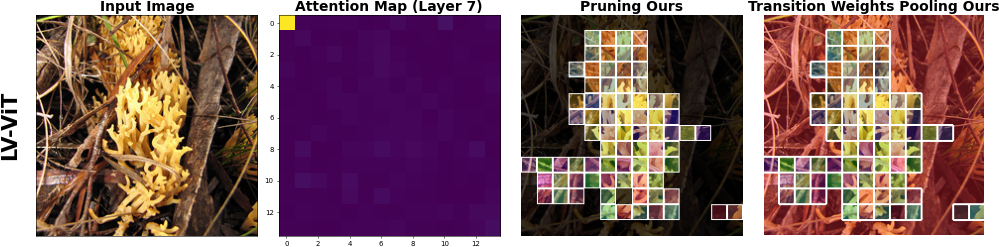} 
    \end{subfigure}
    \hfill
    \begin{subfigure}[b]{0.48\linewidth}
        \centering
        \includegraphics[width=\linewidth]{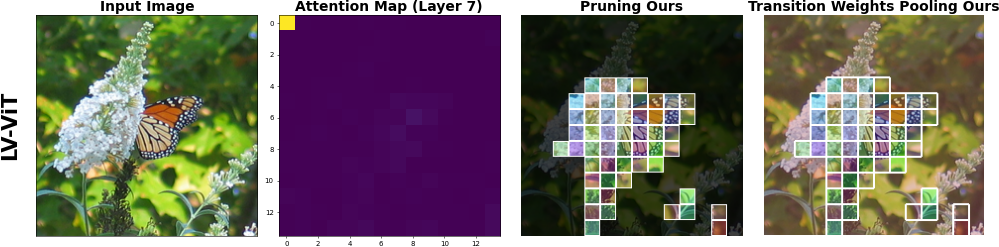}
    \end{subfigure}
    \begin{subfigure}[b]{0.48\linewidth}
        \centering
        \includegraphics[width=\linewidth]{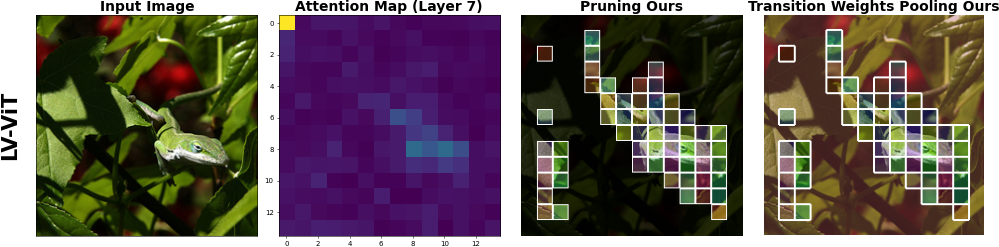} 
    \end{subfigure}
    \hfill
    \begin{subfigure}[b]{0.48\linewidth}
        \centering
        \includegraphics[width=\linewidth]{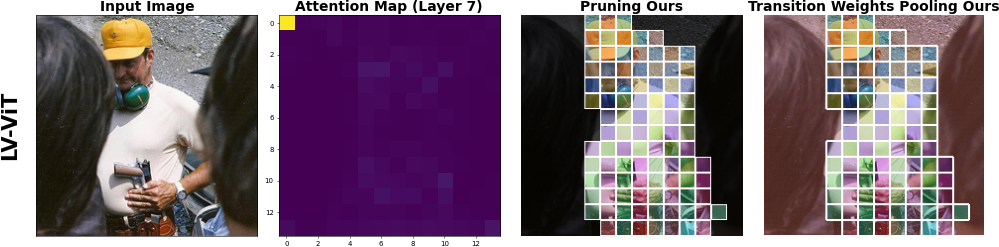}
    \end{subfigure}
    % \begin{subfigure}[b]{0.48\linewidth}
    %     \centering
    %     \includegraphics[width=\linewidth]{appendix/AugReg_appendix_figure/appendix_augreg_mogwa.png} 
    % \end{subfigure}
    % \hfill
    % \begin{subfigure}[b]{0.48\linewidth}
    %     \centering
    %     \includegraphics[width=\linewidth]{appendix/AugReg_appendix_figure/appendix_augreg_monkey.png}
    % \end{subfigure}
    \begin{subfigure}[b]{0.48\linewidth}
        \centering
        \includegraphics[width=\linewidth]{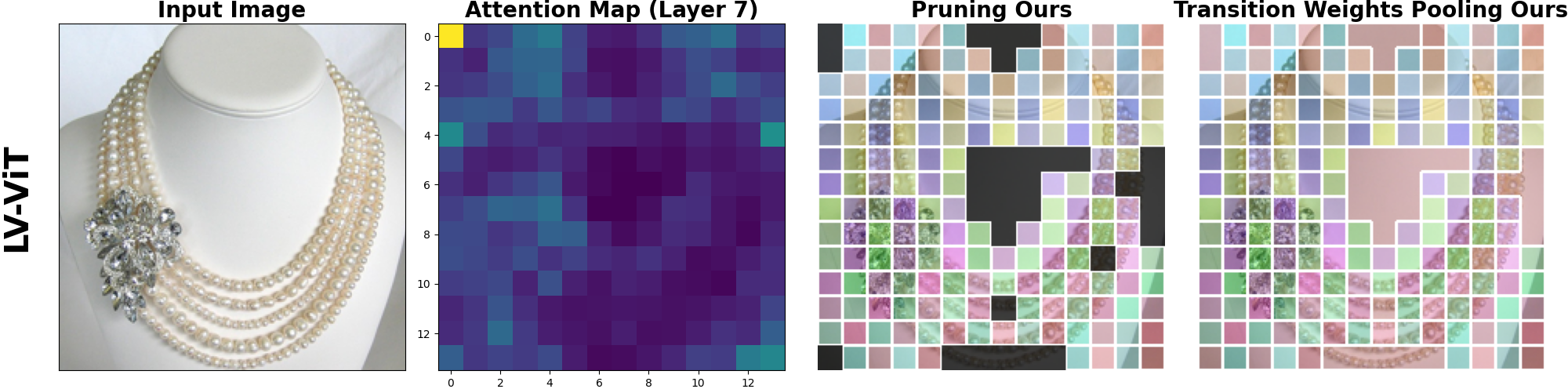} 
    \end{subfigure}
    \hfill
    \begin{subfigure}[b]{0.48\linewidth}
        \centering
        \includegraphics[width=\linewidth]{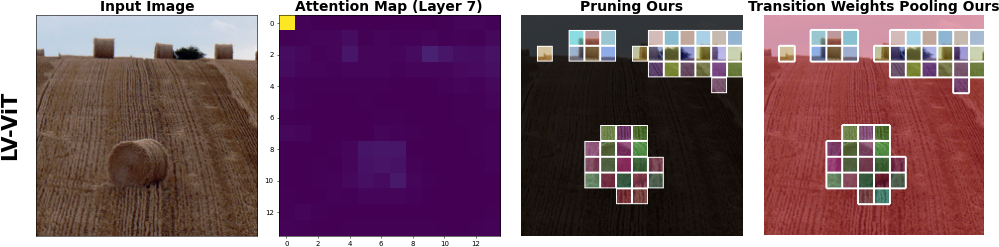}
    \end{subfigure}
    \begin{subfigure}[b]{0.48\linewidth}
        \centering
        \includegraphics[width=\linewidth]{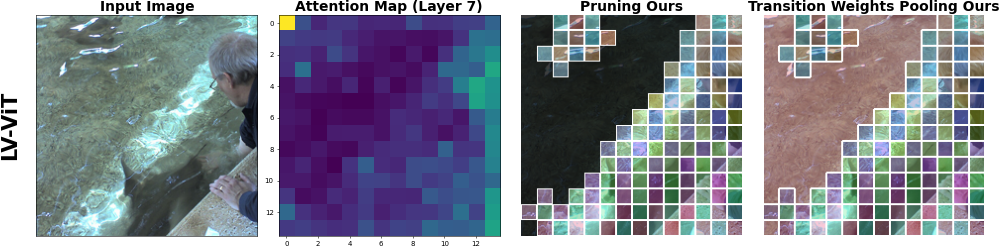} 
    \end{subfigure}
    \hfill
    \begin{subfigure}[b]{0.48\linewidth}
        \centering
        \includegraphics[width=\linewidth]{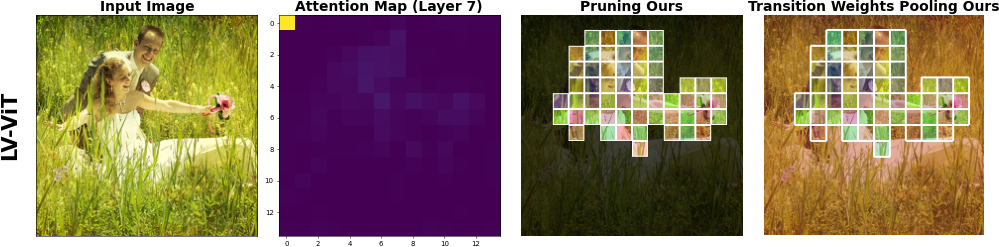}
    \end{subfigure}
    \caption{Qualitative results on LV-ViT-S}
\end{figure}

\section{Random Anchor Qualitative Results}\label{appendix:random_anchor}

\begin{figure}[H]
    \centering
    % === 첫 번째 줄 (Row 1) ===
    \begin{subfigure}[b]{0.48\linewidth}
        \centering
        \includegraphics[width=\linewidth]{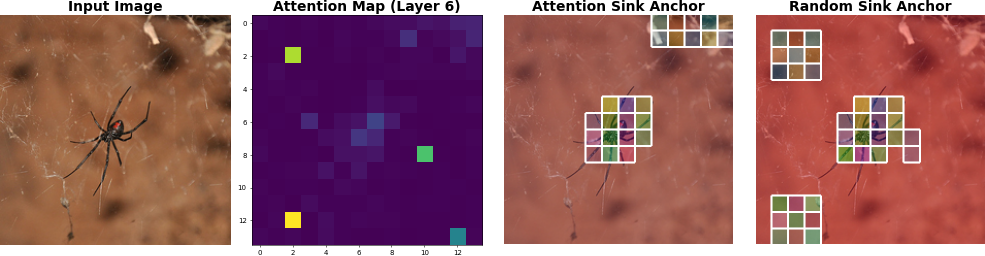}
    \end{subfigure}
    \hfill
    \begin{subfigure}[b]{0.48\linewidth}
        \centering
        \includegraphics[width=\linewidth]{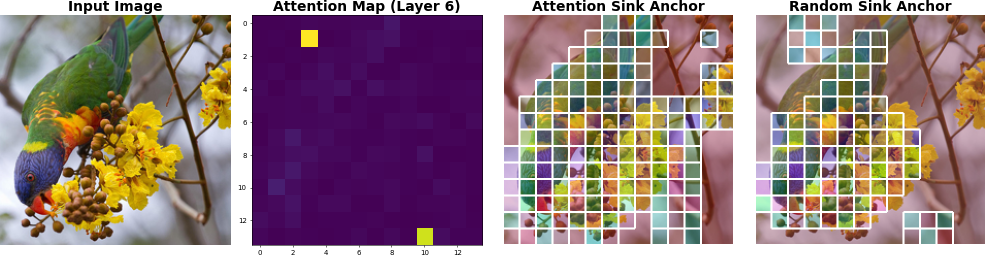}
    \end{subfigure} 
    \begin{subfigure}[b]{0.48\linewidth}
        \centering
        \includegraphics[width=\linewidth]{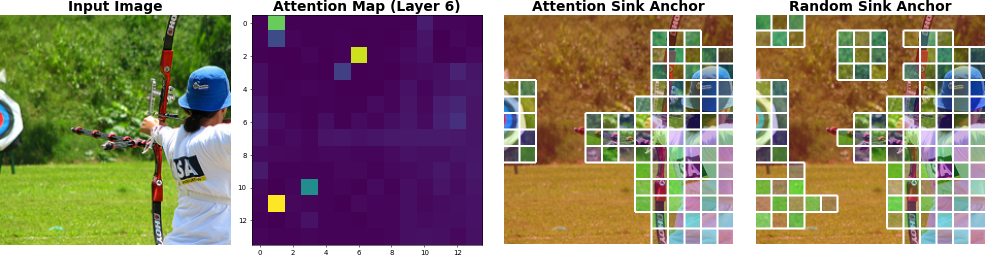} 
    \end{subfigure}
    \hfill
    \begin{subfigure}[b]{0.48\linewidth}
        \centering
        \includegraphics[width=\linewidth]{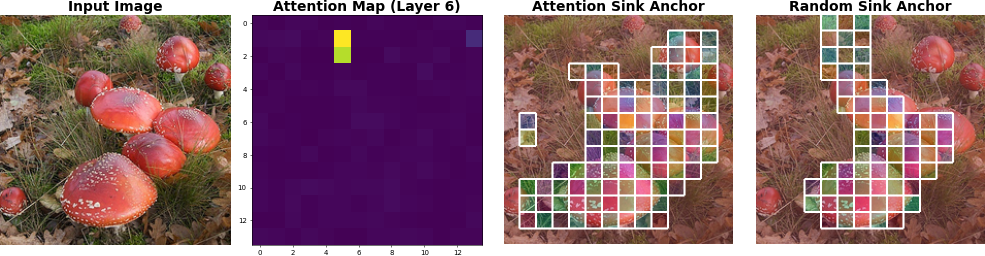}
    \end{subfigure}
    \caption{Positive case of random anchor selection. When the randomly selected anchor fortuitously lands on a semantically neutral background region, it effectively mimics the role of an attention sink. In this case, the diffusion distance-based clustering successfully preserves the semantically meaningful foreground objects.}
    \label{appendix:positive_ra}
\end{figure}

\begin{figure}[H]
    \centering
    % === 첫 번째 줄 (Row 1) ===
    \vspace{-5pt}
    \begin{subfigure}[b]{0.48\linewidth}
        \centering
        \includegraphics[width=\linewidth]{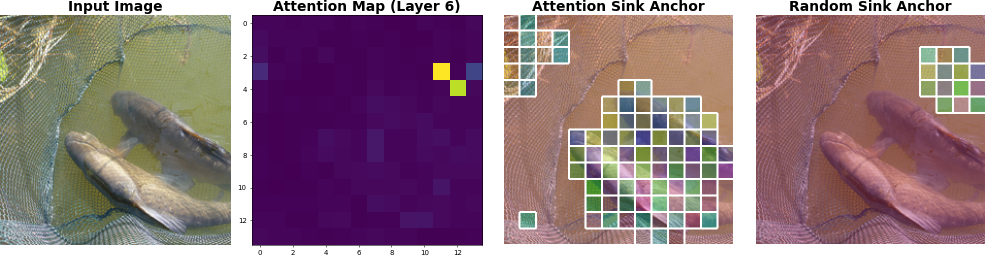}
    \end{subfigure}
    \hfill
    \begin{subfigure}[b]{0.48\linewidth}
        \centering
        \includegraphics[width=\linewidth]{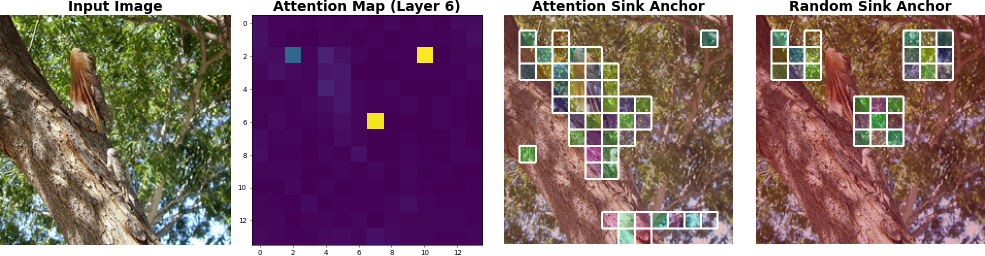}
    \end{subfigure} 
    \begin{subfigure}[b]{0.48\linewidth}
        \centering
        \includegraphics[width=\linewidth]{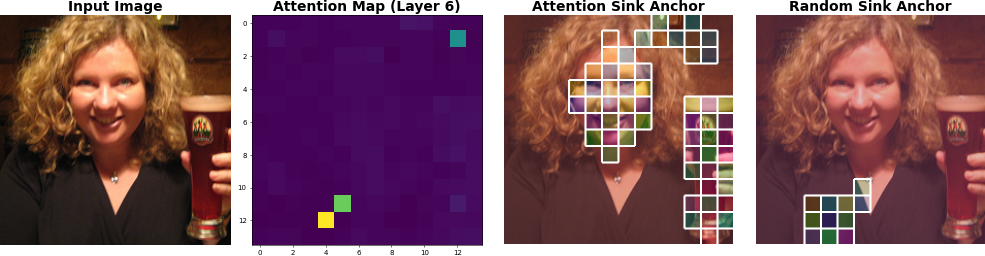} 
    \end{subfigure}
    \hfill
    \begin{subfigure}[b]{0.48\linewidth}
        \centering
        \includegraphics[width=\linewidth]{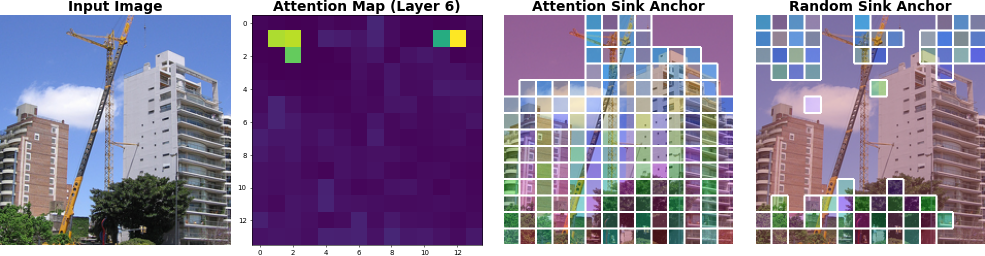}
    \end{subfigure}
    \caption{Negative case of random anchor selection. When the anchor is inadvertently assigned to a key foreground object, the algorithm computes near-zero diffusion distances for the object tokens. This causes the target object to be incorrectly pooled as background, paradoxically preserving irrelevant background patches instead.}
    \label{appendix:negative_ra}
\end{figure}

% sink가 기하학적 anchor로써 중요한 역할을 할 수 있는지에 대한 정량적 실험을 \ref{tab:anchor}에서 진행하였다. 정성적인 결과를 보기 위한 결과는 다음과 같다. \ref{appendix:positive_ra}, \ref{appendix:negative_ra}. 이 결과에서 anchor가 운 좋게 배경을 잡고 그 배경을 바탕으로 diffusion distance를 구하였다면 \ref{appendix:positive_ra}에서 볼 수 있듯 그래도 안정적인 결과를 보여줄 수 있다. 하지만 anchor가 중요한 객체를 잡는 경우 \ref{appendix:negative_ra}에서 볼 수 있듯 중요한 객체 정보가 망가지며 엉뚱한 image token 만을 살리는 모습을 볼 수 있다.
The fact that the attention sink acts as a crucial geometric anchor for semantic separation in our framework is clearly demonstrated by the quantitative results in Table~\ref{tab:anchor}. Qualitative analyses of using a random token as the anchor are presented in Figure~\ref{appendix:positive_ra} and Figure~\ref{appendix:negative_ra}. When a randomly selected anchor fortuitously falls within a semantically meaningless background region (Figure~\ref{appendix:positive_ra}), it assumes a role similar to the sink, enabling diffusion distance-based clustering to relatively stably isolate the foreground. However, a critical failure occurs when the random anchor is assigned to a key foreground object (Figure~\ref{appendix:negative_ra}). In this scenario, the diffusion distances between the tokens constituting the primary object and the anchor approach zero. Consequently, the algorithm misidentifies these foreground tokens as redundant background and prioritizes them for compression (pooling). As a result, critical object information is severely corrupted, while semantically irrelevant background tokens are paradoxically retained. These findings strongly emphasize that dynamically detecting a semantically neutral attention sink to serve as the geometric anchor is an indispensable component of our clustering framework.

% \section{Technical appendices and supplementary material}
% Technical appendices with additional results, figures, graphs, and proofs may be submitted with the paper submission before the full submission deadline (see above). You can upload a ZIP file for videos or code, but do not upload a separate PDF file for the appendix. There is no page limit for the technical appendices. 

% Note: Think of the appendix as ``optional reading'' for reviewers. The paper must be able to stand alone without the appendix; for example, adding critical experiments that support the main claims to an appendix is inappropriate. 

%%%%%%%%%%%%%%%%%%%%%%%%%%%%%%%%%%%%%%%%%%%%%%%%%%%%%%%%%%%%

\newpage
\section*{NeurIPS Paper Checklist}

\begin{enumerate}

\item {\bf Claims}
    \item[] Question: Do the main claims made in the abstract and introduction accurately reflect the paper's contributions and scope?
    \item[] Answer: \answerYes{} % Replace by \answerYes{}, \answerNo{}, or \answerNA{}.
    \item[] Justification: The main claims made in the abstract and introduction, specifically regarding the limitations of local metrics and the effectiveness of our global Markov chain-based geometric clustering, are strictly evaluated and supported by extensive experiments in Section 4.

\item {\bf Limitations}
    \item[] Question: Does the paper discuss the limitations of the work performed by the authors?
    \item[] Answer: \answerYes{} % Replace by \answerYes{}, \answerNo{}, or \answerNA{}.
    \item[] Justification: We dedicate a specific paragraph in Section 5 (Discussion) to explicitly discuss the limitations of our work, including assumptions about semantic neutrality of the sink, applicability to non-patch-based architectures, and the static cluster count.

\item {\bf Theory assumptions and proofs}
    \item[] Question: For each theoretical result, does the paper provide the full set of assumptions and a complete (and correct) proof?
    \item[] Answer: \answerYes{} % Replace by \answerYes{}, \answerNo{}, or \answerNA{}.
    \item[] Justification: The theoretical foundations of our method, including the proposition of the attention sink as a mass-accumulating attractor and the properties of diffusion distance, are formally defined and proven in Appendices A and B.

    \item {\bf Experimental result reproducibility}
    \item[] Question: Does the paper fully disclose all the information needed to reproduce the main experimental results of the paper to the extent that it affects the main claims and/or conclusions of the paper (regardless of whether the code and data are provided or not)?
    \item[] Answer: \answerYes{} % Replace by \answerYes{}, \answerNo{}, or \answerNA{}.
    \item[] Justification: We detail all necessary experimental settings, baseline comparisons, and dataset information in Section 4. Further hyperparameter sensitivities and configurations are provided in Appendices I, J and K to ensure full reproducibility.

\item {\bf Open access to data and code}
    \item[] Question: Does the paper provide open access to the data and code, with sufficient instructions to faithfully reproduce the main experimental results, as described in supplemental material?
    \item[] Answer: \answerYes{} % Replace by \answerYes{}, \answerNo{}, or \answerNA{}.
    \item[] Justification: We provide the complete source code, a representative subset of data samples, and detailed instructions required to faithfully reproduce the main experimental results in the supplementary material.

\item {\bf Experimental setting/details}
    \item[] Question: Does the paper specify all the training and test details (e.g., data splits, hyperparameters, how they were chosen, type of optimizer) necessary to understand the results?
    \item[] Answer: \answerYes{} % Replace by \answerYes{}, \answerNo{}, or \answerNA{}.
    \item[] Justification: Detailed information regarding the evaluated models (DeiT, ViT-AugReg, LV-ViT, LLaVA), datasets (ImageNet-1K, Kinetics-400, multiple VLM benchmarks), and specific hyperparameter choices ($K$, $\tau$, $\alpha$) are fully disclosed in Section 4 and the Appendices.

\item {\bf Experiment statistical significance}
    \item[] Question: Does the paper report error bars suitably and correctly defined or other appropriate information about the statistical significance of the experiments?
    \item[] Answer: \answerNA{} % Replace by \answerYes{}, \answerNo{}, or \answerNA{}.
    \item[] Justification: Our proposed ASAP framework is an entirely training-free, deterministic inference algorithm. Given a fixed pre-trained model and dataset, the outputs and evaluation metrics do not suffer from the statistical variance typically associated with stochastic model training.

\item {\bf Experiments compute resources}
    \item[] Question: For each experiment, does the paper provide sufficient information on the computer resources (type of compute workers, memory, time of execution) needed to reproduce the experiments?
    \item[] Answer: \answerYes{} % Replace by \answerYes{}, \answerNo{}, or \answerNA{}.
    \item[] Justification: We explicitly specify the hardware infrastructure (a single RTX 4090 GPU for vision tasks and an A100 GPU for VLMs) in the introduction of Section 4. Additionally, detailed computational costs (FLOPs, Throughput) are thoroughly analyzed in Section 4 and Appendix F.
    
\item {\bf Code of ethics}
    \item[] Question: Does the research conducted in the paper conform, in every respect, with the NeurIPS Code of Ethics \url{https://neurips.cc/public/EthicsGuidelines}?
    \item[] Answer: \answerYes{} % Replace by \answerYes{}, \answerNo{}, or \answerNA{}.
    \item[] Justification: The research involves optimizing inference efficiency for existing pre-trained models using standard benchmark datasets and strictly conforms to the NeurIPS Code of Ethics.

\item {\bf Broader impacts}
    \item[] Question: Does the paper discuss both potential positive societal impacts and negative societal impacts of the work performed?
    \item[] Answer: \answerNo{} % Replace by \answerYes{}, \answerNo{}, or \answerNA{}.
    \item[] Justification: This work focuses on foundational algorithmic improvements for inference acceleration and computational efficiency in Vision Transformers. It does not introduce direct or specific negative societal impacts that warrant a dedicated discussion beyond general AI efficiency.
    
\item {\bf Safeguards}
    \item[] Question: Does the paper describe safeguards that have been put in place for responsible release of data or models that have a high risk for misuse (e.g., pre-trained language models, image generators, or scraped datasets)?
    \item[] Answer: \answerNA{} % Replace by \answerYes{}, \answerNo{}, or \answerNA{}.
    \item[] Justification: The paper proposes an inference-time token reduction algorithm and does not release new high-risk foundational models, pre-trained language models, or scraped datasets.

\item {\bf Licenses for existing assets}
    \item[] Question: Are the creators or original owners of assets (e.g., code, data, models), used in the paper, properly credited and are the license and terms of use explicitly mentioned and properly respected?
    \item[] Answer: \answerYes{} % Replace by \answerYes{}, \answerNo{}, or \answerNA{}.
    \item[] Justification: All existing assets, including standard public datasets (e.g., ImageNet, Kinetics-400, VQA v2, GQA) and pre-trained model architectures, are properly cited and referenced in the References section.

\item {\bf New assets}
    \item[] Question: Are new assets introduced in the paper well documented and is the documentation provided alongside the assets?
    \item[] Answer: \answerNA{} % Replace by \answerYes{}, \answerNo{}, or \answerNA{}.
    \item[] Justification: The paper does not introduce or release any new datasets or novel pre-trained model weights.

\item {\bf Crowdsourcing and research with human subjects}
    \item[] Question: For crowdsourcing experiments and research with human subjects, does the paper include the full text of instructions given to participants and screenshots, if applicable, as well as details about compensation (if any)? 
    \item[] Answer: \answerNA{} % Replace by \answerYes{}, \answerNo{}, or \answerNA{}.
    \item[] Justification: This research is purely computational and does not involve crowdsourcing or human subjects.%\justificationTODO{}

\item {\bf Institutional review board (IRB) approvals or equivalent for research with human subjects}
    \item[] Question: Does the paper describe potential risks incurred by study participants, whether such risks were disclosed to the subjects, and whether Institutional Review Board (IRB) approvals (or an equivalent approval/review based on the requirements of your country or institution) were obtained?
    \item[] Answer: \answerNA{} % Replace by \answerYes{}, \answerNo{}, or \answerNA{}.
    \item[] Justification: This research does not involve human subjects.%\justificationTODO{}

\item {\bf Declaration of LLM usage}
    \item[] Question: Does the paper describe the usage of LLMs if it is an important, original, or non-standard component of the core methods in this research? Note that if the LLM is used only for writing, editing, or formatting purposes and does \emph{not} impact the core methodology, scientific rigor, or originality of the research, declaration is not required.
    %this research? 
    \item[] Answer: \answerNA{} % Replace by \answerYes{}, \answerNo{}, or \answerNA{}.
    \item[] Justification: The core methodological development of our token reduction framework does not involve LLMs as an important, original, or non-standard component.

\end{enumerate}

\end{document}